%% file: main.tex
\documentclass[11pt]{article}

\input{macros}

\title{\sys: Diffusion Language Models for Agentic Planning}
\author{\begin{tabular}{@{}ccc@{}}
Syamantak Kumar\thanks{University of Texas at Austin. Correspondence: \texttt{syamantak@utexas.edu}.}
& Jiang Guo\thanks{Amazon. Correspondence: \texttt{gujiang@amazon.com}.}
& Hassan Hamad\footnotemark[2] \\[0.5em]
Hideo Kobayashi\footnotemark[2]
& Yi Xiang\footnotemark[2]
& Yezhou Yang\footnotemark[2] \\[0.5em]
Yanjun Qi\footnotemark[2]
& Daniele Bonadiman\footnotemark[2]
& Jiarong Jiang\footnotemark[2]
\end{tabular}}
\date{}

\begin{document}
\allowdisplaybreaks
\raggedbottom
\pagenumbering{gobble}
\maketitle
\input{abstract}

\thispagestyle{empty}

\clearpage
\tableofcontents
\thispagestyle{empty}
\clearpage
\pagenumbering{arabic}
\suppressfloats[t]

\input{introduction}

\input{relatedwork}
\input{problemsetup}
\input{method}
\input{experiments}
\input{conclusion}

\clearpage
\bibliographystyle{plainnat}
\bibliography{refs}

\clearpage
\appendix
\setcounter{table}{0}
\setcounter{figure}{0}
\renewcommand{\thetable}{A.\arabic{table}}
\renewcommand{\thefigure}{A.\arabic{figure}}
\input{appendix}
\end{document}

%% file: macros.tex
\usepackage{fullpage}
\usepackage{parskip}
\usepackage{microtype}
\usepackage[utf8]{inputenc}
\usepackage{ifthen}
\usepackage[authoryear,round]{natbib}
\input{paper_macros}
\usepackage{lmodern}
\input{t1lmss.fd}
\input{t1lmr.fd}
\DeclareFontShape{T1}{lmr}{bx}{sc}{<->ssub*lmr/m/sc}{}
\DeclareFontShape{T1}{lmss}{m}{sc}{<->ssub*lmr/m/sc}{}
\DeclareFontShape{T1}{lmss}{bx}{sc}{<->ssub*lmr/m/sc}{}

\providecommand{\pr}[2]{
    \ifthenelse{\equal{#1}{}}{
        \mathbb{P}\left[ #2 \right]
    }
    {
        \underset{#1}{\mathbb{P}}\left[ #2 \right]
    }
}
\providecommand{\ex}[2]{
    \ifthenelse{\equal{#1}{}}{
        \mathbb{E}\left[ #2 \right]
    }
    {
        \underset{#1}{\mathbb{E}}\left[ #2 \right]
    }
}
\providecommand{\var}[2]{
    \ifthenelse{\equal{#1}{}}{
        \mathrm{Var}\left( #2 \right)
    }
    {
        \underset{#1}{\mathrm{Var}}\left( #2 \right)
    }
}

\providecommand{\1}{\mathbf{1}}



%% file: paper_macros.tex
\usepackage[T1]{fontenc}
\usepackage{amsmath,amssymb,amsthm,mathtools}
\usepackage{bm,bbm,bbold}
\usepackage{accents}
\usepackage{booktabs,array,multirow}
\usepackage{graphicx}
\usepackage{placeins}
\usepackage{subcaption}
\graphicspath{{./figs/}}
\usepackage{xcolor}
\usepackage[english]{babel}
\usepackage{enumitem}
\usepackage{algorithm}
\usepackage{algpseudocode}
\usepackage{tikz}
\usepackage{pgfplots}
\pgfplotsset{compat=1.18}
\usetikzlibrary{arrows.meta,positioning,calc,fit,backgrounds,shapes.geometric,shapes.misc}
\usepackage{hyperref}
\hypersetup{
	colorlinks=true,
	linkcolor=blue!70!black,
	citecolor=blue!70!black,
	urlcolor=blue!70!black
}
\usepackage[capitalise,noabbrev]{cleveref}
\usepackage{thm-restate}

\usepackage{xspace}
\newcommand{\sys}{\textsc{Plan-and-Patch}\xspace}

\usepackage{diagbox}

\newcommand{\diffp}{\textsc{Diff-Plan}\xspace}   
\newcommand{\arp}{\textsc{AR-Plan}\xspace}        
\newcommand{\repairregion}{R}
\newcommand{\repairfeedback}{f_{\repairregion}}
\newcommand{\plan}{\tau}
\newcommand{\planvocab}{\mathcal{V}}
\newcommand{\planlength}{L}
\newcommand{\taskquery}{Q}
\newcommand{\targetseq}{A}
\newcommand{\targetlength}{T}
\newcommand{\modelparams}{\theta}
\newcommand{\tokenmodel}{p_{\modelparams}}
\newcommand{\maskrate}{t}
\newcommand{\maskpositions}{M_{\maskrate}}
\newcommand{\corruptedtarget}{\widetilde{\targetseq}_{\maskrate}}

\newcommand{\exechistory}{h}
\newcommand{\replacement}{\rho}

\newcommand{\plantool}{u}
\newcommand{\toolargs}{a}

\newcommand{\blocksize}{b}
\newcommand{\leafcount}{H}
\newcommand{\repaircapacity}{K_0}

\newcommand{\supportthr}{\kappa}
\newcommand{\supportradius}{r}
\definecolor{diffcol}{RGB}{36,111,159}
\definecolor{arcol}{RGB}{203,112,54}
\input{figs/visual_style.tex}

\newtheorem{definition}{Definition}

\newcommand{\1}{\mathbbm{1}}

\newcommand{\id}{\mathbf{I}}

\usepackage{xcolor}
\definecolor{burntorange}{rgb}{0.8, 0.33, 0.0}

\usepackage{xcolor}

\newcommand{\confthr}{\eta}
\newcommand{\confmin}{\eta_{\min}}

%% file: figs/visual_style.tex
\definecolor{ink}{RGB}{31,49,68}
\definecolor{muted}{RGB}{89,104,118}
\definecolor{rulecol}{RGB}{211,220,227}
\definecolor{panelbg}{RGB}{244,247,250}
\definecolor{repaircol}{RGB}{29,120,112}
\definecolor{errorcol}{RGB}{161,64,68}
\usetikzlibrary{patterns.meta}
\tikzset{
  fp bar/.style={solid,line width=.35pt},
  fp diff bar/.style={fp bar,fill=diffcol!30,draw=diffcol!75,
    postaction={pattern={Lines[angle=45,distance=3pt,line width=.2pt]},pattern color=diffcol!75}},
  fp ar bar/.style={fp bar,fill=arcol!30,draw=arcol!75,
    postaction={pattern={Lines[angle=-45,distance=3pt,line width=.2pt]},pattern color=arcol!75}},
  fp pooled bar/.style={fp bar,fill=muted!25,draw=muted!75},
  fp baseline bar/.style={fp bar,fill=muted!40,draw=muted!75},
  fp text/.style={font=\sffamily\fontsize{7.5}{8.6}\selectfont,text=ink},
  fp title/.style={font=\sffamily\bfseries\fontsize{8.2}{9.4}\selectfont,text=ink,anchor=west,inner sep=0pt},
  fp label/.style={font=\sffamily\fontsize{7.0}{8}\selectfont,text=muted,inner sep=1pt},
  fp value/.style={font=\sffamily\bfseries\fontsize{7.2}{8.2}\selectfont,text=ink,inner sep=1pt},
  fp grid/.style={draw=rulecol,line width=0.3pt},
  fp axis/.style={draw=muted,line width=0.4pt},
  fp arrow/.style={-{Stealth[length=1.5mm,width=1.1mm]},draw=muted,line width=0.65pt},
  fp card/.style={draw=rulecol,fill=panelbg,line width=0.55pt,rounded corners=1.4pt,
    align=left,inner sep=5pt,font=\sffamily\fontsize{7.2}{8.4}\selectfont,text=ink}
}

%% file: abstract.tex
\begin{abstract}
Planning is increasingly important for long-horizon agents, where successful
execution requires coordinating subgoals, tool use, and intermediate outcomes
over many steps. Yet assumptions made during planning may be invalidated by the
environment, tools may return unexpected results, or actions may fail. Effective
agents must therefore not only generate plans, but also revise them. Such
revisions often affect only part of a plan, leaving the preceding and subsequent
structure intact. Rather than regenerating the entire plan and risking unnecessary
changes, repair can regenerate the affected region conditioned on the preserved
prefix and suffix.
We introduce \sys, a plan-and-act framework in which a diffusion
language model (dLLM) generates
a structured, program-like plan through parallel unmasking and repairs it by
filling in selected regions while keeping the surrounding steps fixed.
We compare DreamReasoner-8B and Qwen3-8B as diffusion and autoregressive (AR)
planners. On Natural Plan without task-specific training, diffusion ($53.7\%$)
achieves nearly twice the plan repair success rate of AR ($27.0\%$).
After task-specific training on agentic benchmarks ALFWorld
and TextCraft, the planners achieve similar
observed success in plan generation, while diffusion
reduces mean plan-generation latency by $39$--$46\%$ relative to AR.
Our results show that \sys provides a framework for faster plan generation
and effective plan repair in long-horizon agents.
\end{abstract}

%% file: introduction.tex
\section{Introduction}
\label{sec:intro}

Planning helps long-horizon agents coordinate subgoals, tool use, and intermediate
outcomes. Plan-and-act methods first construct a multi-step plan, then turn
its steps into environment actions \citep{plansolve,adaplanner,planandact}. This separation
can improve task success over workflows without an explicit planning stage
\citep{adapt,adaplanner,plannermatters,hiper,recap}. Program-like representations
make the plan's structure explicit through subgoals and executable steps
\citep{progprompt,codeaspolicies,voyager,deps,recap}. Yet assumptions made during
planning may be invalidated by the environment, tools may return unexpected
results, or actions may fail. Such failures can arise because the planner has
incomplete knowledge of the environment, misjudges action preconditions, or
relies on conditions that change during execution \citep{deps,adaplanner,planandact}.
Effective agents must therefore both generate
plans and revise them in response to environment feedback.

Planning requires reasoning about dependencies among actions and their long-term
consequences for achieving the goal \citep{pitfalls}.
LLM planners address this challenge by anticipating future states \citep{rap}
and decomposing complex tasks into simpler subgoals \citep{adapt,progprompt}.
Plan format also affects the performance of an agent significantly. \citet{planahead} compare sequential subgoals,
narrative, pseudocode, and checklists, finding that web-agent performance
depends on both representation and planner--executor pairing.
We propose a \textbf{structured plan representation} with explicit subgoals and
executable steps to provide meaningful units for revision. A replacement must
remain compatible with the preceding and subsequent plan context.

Such revisions often affect only part of a plan. However, existing
plan-and-act agents commonly revise globally \citep{adapt,planandact,reflexion}.
In this work, we model \textbf{plan repair as an infilling problem}, identifying the affected region
and regenerating it conditioned on the preserved prefix and suffix.

\input{figs/fig_framework.tex}

While autoregressive (AR) models can be adapted to infilling through explicit
fill-in-the-middle training objectives \citep{fim,incoder}, masked diffusion
language models (dLLMs) naturally learn to reconstruct masked regions
conditioned on their surrounding context \citep{d3pm,sedd,mdlm}.
Recent work demonstrates competitive code generation \citep{codefusion,diffucoder}
and code infilling \citep{dreamon,mercury} with diffusion language models,
including higher generation throughput at comparable coding performance
\citep{mercury}. These results motivate exploring diffusion models for structured
plan generation and repair. We therefore ask,

\emph{Can we design an agentic planning framework that exploits diffusion language
models' native capabilities for parallel generation and infilling?}

To address this question, we introduce \sys (\Cref{fig:framework}), a
plan-and-act framework that uses a single dLLM trained for both plan generation
and repair. The planner generates a complete structured plan through parallel
unmasking, and a separate executor translates its steps into environment
actions. If execution fails, the executor also serves as the failure detector,
using the current plan, execution history, and environment feedback to select
a repair region and explain the required revision. The planner uses this repair
feedback to infill the region while preserving the surrounding plan text.

We evaluate diffusion and AR planners on ALFWorld for household task execution
\citep{alfworld} and TextCraft for dependency-based crafting \citep{adapt},
using task-specific training, a shared structured plan representation, and a
common executor. The planners achieve similar observed performance in plan
generation and repair, with a slight edge to AR, while diffusion reduces mean
plan-generation latency by $39$--$46\%$.
On Natural Plan for travel and scheduling \citep{naturalplan} without
task-specific training, diffusion ($53.7\%$) nearly doubles the plan repair
success rate of AR ($27.0\%$). Our main contributions are as follows.

\begin{enumerate}
    \item We introduce \sys, a plan-and-act framework that jointly trains a single
diffusion language model for complete-plan generation and plan repair,
modeling repair as infilling conditioned on surrounding plan context and
repair feedback.
\item We develop a data construction pipeline that combines synthetic repair
examples, generated through structured perturbations of verified plans, with
naturally occurring execution failures. Together with successful trajectories,
these examples support joint training for generation and repair.
\item We present a training recipe and curriculum for joint plan generation and
repair with diffusion language models, using objective-specific block-size
sampling and decoding. We describe the training choices that support joint
generation and repair, and analyze how plan representation, executor choice,
and repair-region selection affect performance.
\end{enumerate}

%% file: figs/fig_framework.tex
\begin{figure}[t]
\centering
\resizebox{0.84\linewidth}{!}{\input{figure_assets/framework.tikz}}
\caption{\textbf{The \sys framework.}
The same planner generates plans and infills selected regions. After an
execution failure, the executor acts as the failure detector, selecting a region using execution history
and environment feedback. The planner infills it conditioned on repair feedback,
with the prefix and suffix fixed.
\Cref{fig:teaser} gives a worked example.}
\label{fig:framework}
\end{figure}
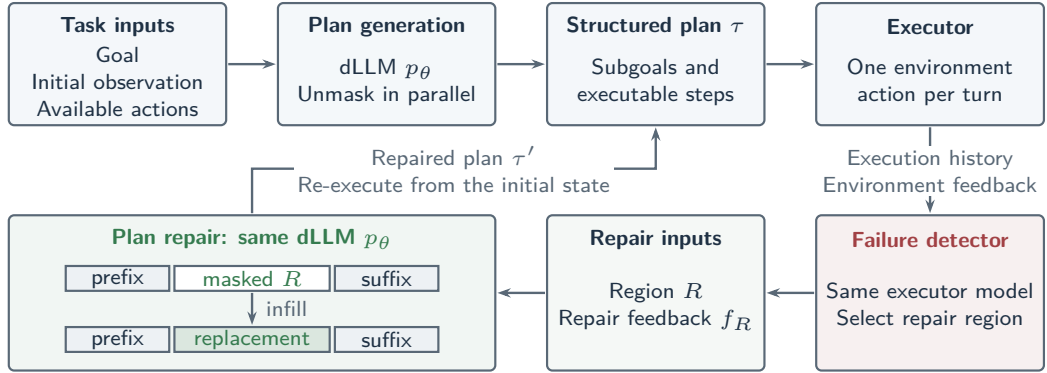

%% file: figure_assets/framework.tikz
\begin{tikzpicture}[x=1cm,y=1cm,text=ink,
 font=\sffamily\fontsize{6.0}{7.4}\selectfont,
 ov border/.style={draw=muted,solid,line width=.6pt,draw opacity=1},
 ov box/.style={ov border,fill=panelbg,rounded corners=1.5pt},
 ov planner/.style={ov box,fill=diffcol!5},
 ov title/.style={inner sep=0pt,font=\sffamily\bfseries\fontsize{6.0}{7.4}\selectfont},
 ov note/.style={inner sep=0pt,text=muted},
 ov arrow/.style={-{Stealth[length=1.3mm,width=.9mm]},draw=muted,line width=.65pt}]
\definecolor{ovrepair}{RGB}{53,123,79}
\path[use as bounding box] (0,.02) rectangle (10,-3.55);

\path[ov box] (0,0) rectangle (2.10,-1.18);
\node[ov title] at (1.05,-.23) {Task inputs};
\node[align=center,inner sep=0pt] at (1.05,-.76)
 {Goal\\Initial observation\\Available actions};
\draw[ov arrow] (2.12,-.59)--(2.58,-.59);

\path[ov planner] (2.60,0) rectangle (4.70,-1.18);
\node[ov title] at (3.65,-.23) {Plan generation};
\node[align=center,inner sep=0pt] at (3.65,-.76)
 {dLLM $\tokenmodel$\\Unmask in parallel};
\draw[ov arrow] (4.72,-.59)--(5.18,-.59);

\path[ov box] (5.20,0) rectangle (7.30,-1.18);
\node[ov title] at (6.25,-.23) {Structured plan $\plan$};
\node[align=center,inner sep=0pt] at (6.25,-.76)
 {Subgoals and\\executable steps};
\draw[ov arrow] (7.32,-.59)--(7.78,-.59);

\path[ov box] (7.80,0) rectangle (10,-1.18);
\node[ov title] at (8.90,-.23) {Executor};
\node[align=center,inner sep=0pt] at (8.90,-.76)
 {One environment\\action per turn};
\draw[ov arrow] (8.90,-1.20)--(8.90,-2.03);
\node[ov note,align=center,fill=white,inner sep=1pt] at (8.90,-1.62)
 {Execution history\\Environment feedback};

\path[ov box,fill=errorcol!8] (7.80,-2.05) rectangle (10,-3.55);
\node[ov title,text=errorcol] at (8.90,-2.28) {Failure detector};
\node[align=center,inner sep=0pt] at (8.90,-2.93)
 {Same executor model\\Select repair region};
\draw[ov arrow] (7.78,-2.80)--(7.32,-2.80);

\path[ov box,fill=ovrepair!5] (5.20,-2.05) rectangle (7.30,-3.55);
\node[ov title] at (6.25,-2.28) {Repair inputs};
\node[align=center,inner sep=0pt] at (6.25,-2.93)
 {Region $\repairregion$\\Repair feedback $\repairfeedback$};
\draw[ov arrow] (5.18,-2.80)--(4.72,-2.80);

\path[ov box,fill=ovrepair!7] (0,-2.05) rectangle (4.70,-3.55);
\node[ov title,text=ovrepair] at (2.35,-2.28) {Plan repair: same dLLM $\tokenmodel$};
\foreach \yy in {2.54,3.13} {
 \path[ov border,fill=rulecol!40]
  (.55,-\yy) rectangle (1.55,-\yy-.24);
 \path[ov border,fill=rulecol!40]
  (3.15,-\yy) rectangle (4.15,-\yy-.24);
 \node[inner sep=0pt]
  at (1.05,-\yy-.12) {prefix};
 \node[inner sep=0pt]
  at (3.65,-\yy-.12) {suffix};
}
\path[ov border,fill=white]
 (1.60,-2.54) rectangle (3.10,-2.78);
\node[inner sep=0pt,text=ovrepair]
 at (2.35,-2.66) {masked $\repairregion$};
\draw[ov arrow] (2.35,-2.80)--(2.35,-3.11);
\node[ov note,anchor=west] at (2.48,-2.95) {infill};
\path[ov border,fill=ovrepair!18]
 (1.60,-3.13) rectangle (3.10,-3.37);
\node[inner sep=0pt,text=ovrepair]
 at (2.35,-3.25) {replacement};

\draw[ov arrow] (2.35,-2.03)--(2.35,-1.60)--(6.25,-1.60)--(6.25,-1.20);
\node[ov note,align=center,fill=white,inner sep=1pt] at (4.30,-1.60)
 {Repaired plan $\plan'$\\Re-execute from the initial state};
\end{tikzpicture}

%% file: relatedwork.tex
\section{Related Work}
\label{sec:related}
\vspace{-5pt}

\paragraph{Diffusion language models and agents.}
Masked diffusion language models recover masked tokens from surrounding context,
enabling flexible generation order and bidirectional infilling
\citep{d3pm,sedd,mdlm}. Building on this flexibility, advances in scaling,
post-training, and decoding have improved generation quality, reasoning, and
inference efficiency
\citep{llada,dream7b,llada15,d1,tracerl,fastdllm,didiff}.
Their practical efficiency depends on context length, batching, and the number
of sampling steps \citep{dllmperf}. These models have also been applied to
agents, both as world models that predict environment responses
\citep{mdlmworld} and as ReAct-style agents that interleave action generation
with environment observations
\citep{react,dllmagent,dllmsearcher,llada22}.
In these settings, \citet{diffuagent} identify two systematic failure modes,
``retry loops'' in embodied agents and ``imprecise tool-call formats'' in
tool-calling agents.
Agent-specific training improves reliability and efficiency
\citep{dllmsearcher,llada22}. Under matched fine-tuning, \citet{dllmagent}
also compare diffusion and AR models as agents. We study these models in a
plan-and-act setting, focusing on complete-plan generation and plan repair
through infilling guided by repair feedback.

\vspace{-5pt}
\paragraph{Plan repair and program repair.}
Plan-and-act agents construct multi-step plans and translate their steps into
environment actions \citep{plansolve,adaplanner,planandact}. When execution
fails, these plans can be revised using environment feedback
\citep{adapt,llmplanner,reflexion,selfrefine,innermonologue,deps,reflect,voyager}.
An important choice is which part of the plan to revise. PIVOT and ContextFlow
preserve a validated prefix and rewrite a suffix
\citep{pivot,contextflow}.
Atomic Task Graph preserves validated graph regions and repairs the affected
subgraph \citep{atg}, while L-ICL localizes the first constraint violation and
supplies a corrective example \citep{licl}. To assess repair, controlled replay
helps distinguish the effect of changing a plan from variation across
executions \citep{repairresample}. Our evaluation holds failed plans and repair
regions fixed across diffusion and AR models, and separately assesses region
selection using execution history.

A similar use of feedback appears in program repair, where test failures guide
fault localization and patch validation \citep{aprsurvey,genprog,alpharepair}.
This setting is also captured by LiveCodeBench, which evaluates code self-repair
given an incorrect program, a failing test case, and execution feedback
\citep{livecodebench}. To replace part of a program while retaining the
surrounding code, AR models can use fill-in-the-middle serialization or
training to condition on both the prefix and suffix \citep{fim,incoder,fimeval}.
Masked diffusion naturally supports this infilling operation and has been
applied to code generation, editing, and repair
\citep{codefusion,diffucoder,ancoder,diffrepair}. To accommodate replacements of
different lengths, diffusion methods use deletion or expansion tokens, or
predict the required span length before denoising \citep{dreamon,pill}.
\sys applies this formulation to plan repair, where repair feedback guides
infilling of selected plan regions while the surrounding text remains fixed.

\vspace{-5pt}
\paragraph{Diffusion models for planning.}
In continuous control, diffusion models generate trajectories and may
replan as the environment evolves \citep{diffuser,decisiondiffuser,replandiff}.
In discrete domains, evaluations on Countdown, Sudoku, and trip planning
demonstrate their ability to generate solutions with dependencies across steps
\citep{mgdm,dream7b}. Studies of graph planning further show how later path
tokens can guide earlier decisions
\citep{asymmetry}. A complementary approach separates planning from execution, with
Latent-DARM connecting a diffusion planner to an AR executor through a latent
communication interface for reasoning tasks \citep{latentdarm}. Our work
combines plan generation and repair in interactive environments, using execution
history for region selection and repair feedback for infilling.

%% file: problemsetup.tex
\section{Preliminaries}
\label{sec:formal}

We introduce the structured plan representation, then review masked diffusion
training and the decoding procedures used by \sys.

\begin{definition}[Structured plan]
\label{def:plan}
A structured plan consists of a goal and an ordered hierarchy of subgoals and
executable steps. Each subgoal contains an ordered list of steps or nested
subgoals. Each leaf step specifies an action, its arguments, and an expected
observation. The plan is serialized as a token sequence
$\plan\in\planvocab^{\planlength}$, where $\planvocab$ is the vocabulary and
$\planlength\ge1$ is its token length. Every subgoal and step has an identifier
that locates its contiguous token span in $\plan$; we call these spans
addressable. Reasoning and intent fields also have explicit delimiters.
\end{definition}

\input{figs/fig_plan_tree.tex}

\paragraph{Masked diffusion language models.}
Representing structured plans as token sequences lets us formulate plan
generation and infilling as conditional text generation. Let $\taskquery$ denote the input context,
$\targetseq=(\targetseq_1,\dots,\targetseq_{\targetlength})$ a target sequence
of $\targetlength$ tokens, and $\tokenmodel$ a token model with parameters
$\modelparams$. An AR model generates tokens from left to right, with
conditional log-likelihood
\[
\log P_{\modelparams}(\targetseq\mid\taskquery)
=\sum_{i=1}^{\targetlength}\log\tokenmodel(\targetseq_i\mid\targetseq_{<i},\taskquery).
\]
A masked diffusion model instead learns to reconstruct masked tokens from
their surrounding context. During training, it draws a mask rate
$\maskrate\sim\mathrm{Unif}(0,1)$ and independently replaces each target token
with \texttt{[MASK]} with probability $\maskrate$. Let $\corruptedtarget$
be the corrupted sequence and $\maskpositions$ the set of masked positions.
The model minimizes the masked-token objective \citep{mdlm,sedd}
\[
\mathcal{L}_{\mathrm{MDM}}(\modelparams)
=\mathbb{E}_{\targetseq,\taskquery,\maskrate,\corruptedtarget}\!\left[
-\frac{1}{\maskrate}\sum_{i\in\maskpositions}
\log\tokenmodel(\targetseq_i\mid\corruptedtarget,\taskquery)\right].
\]
The expectation is over training context--target pairs, the mask rate, and
the resulting corruption.

\paragraph{Decoding schedules.}
\label{sec:decoding}
We use DreamReasoner's confidence-based unmasking procedure
\citep{dreamreasoner}, which commits sufficiently confident token predictions
in parallel while keeping resolved tokens fixed. Generation decodes blocks of
$\blocksize$ positions from left to right using a fixed confidence threshold.
Repair jointly infills the selected spans using DreamReasoner's
RelaxedConfidence procedure, which lowers the threshold when nearby positions
are reliable. Decoding settings and forward-pass statistics appear in
\Cref{app:decode}.

Each selected span is replaced by a padded sequence of masked tokens, called
its canvas segment. Following deletion-token approaches \citep{dreamon}, we end
each replacement before the first \texttt{[delete]} token in its segment,
allowing replacements of different lengths. \Cref{sec:repair} specifies the canvas allocation.
The AR baseline jointly generates all replacements in one left-to-right output
sequence, using fill-in-the-middle (FIM) serialization to condition on the
preserved prefix, suffix, and repair feedback.

%% file: figs/fig_plan_tree.tex
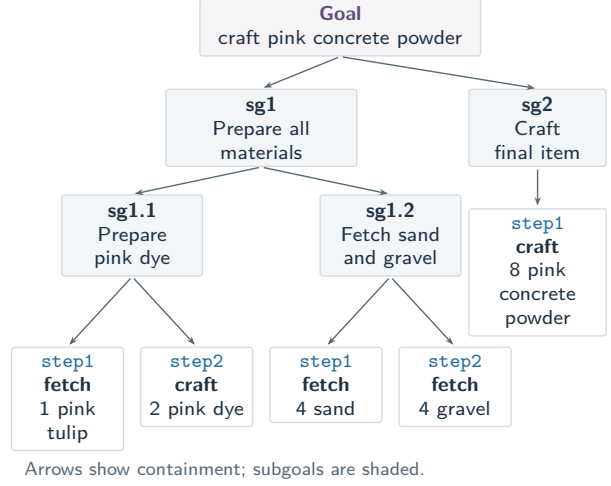
\begin{figure}[t]
\centering
\resizebox{\linewidth}{!}{\input{figure_assets/plan_tree.tikz}}
\caption{\textbf{A TextCraft plan and its ordered hierarchy.}
All five executable steps are shown; reasoning is elided.
Arrows show containment; execution visits leaves in depth-first, left-to-right order.}
\label{fig:plantree}
\end{figure}

%% file: figure_assets/plan_tree.tikz
\begin{tikzpicture}[x=1cm,y=1cm,fp text]
\definecolor{plangoalcol}{RGB}{100,77,125}
\definecolor{planargscol}{RGB}{132,86,43}
\newcommand{\plantag}[1]{{\ttfamily\fontsize{7.0}{8.0}\selectfont\color{ink}#1}}
\newcommand{\plangoaltext}[1]{{\bfseries\color{plangoalcol}#1}}
\newcommand{\plansubgoaltext}[1]{{\color{diffcol!85!black}#1}}
\newcommand{\planargtext}[1]{{\color{planargscol}#1}}
\newcommand{\planobstext}[1]{{\color{ink}#1}}
\node[fp title] at (0,0) {(a) Complete executable plan};
\node[fp card,anchor=north west,fill=panelbg!35,inner sep=6pt,
 text width=8.15cm,font=\sffamily\fontsize{8.0}{8.0}\selectfont,
 align=left] (listing) at (0,-0.35) {%
\plantag{[PLAN\_BEGIN]}\\
\hspace*{0.6em}\plantag{[GOAL]} \plangoaltext{craft pink concrete powder} \plantag{[/GOAL]}\\
\hspace*{0.6em}\plantag{[THINK]} \textellipsis{} \plantag{[/THINK]}\\
\hspace*{0.6em}\plantag{[SG\_BEGIN op=SEQ]} {\bfseries\color{diffcol!85!black}sg1}\\
\hspace*{1.2em}\plantag{[SG\_INTENT]} \plansubgoaltext{prepare all materials needed for pink}\\
\hspace*{1.8em}\plansubgoaltext{concrete powder} \plantag{[/SG\_INTENT]}\\
\hspace*{1.2em}\plantag{[SG\_BEGIN op=SEQ]} {\bfseries\color{diffcol!85!black}sg1.1}\\
\hspace*{1.8em}\plantag{[SG\_INTENT]} \plansubgoaltext{prepare pink dye by fetching pink tulip}\\
\hspace*{2.4em}\plansubgoaltext{and crafting it} \plantag{[/SG\_INTENT]}\\
\hspace*{1.8em}\plantag{[STEP\_BEGIN]} {\bfseries\color{diffcol!85!black}sg1.1.step1}\\
\hspace*{2.4em}\plantag{[ACTION]} \textbf{fetch} \plantag{[/ACTION]} \plantag{[ARGS]} \planargtext{1 pink tulip} \plantag{[/ARGS]}\\
\hspace*{2.4em}\plantag{[EXPECTED\_OBS]} \planobstext{Got 1 pink tulip} \plantag{[/EXPECTED\_OBS]}\\
\hspace*{1.8em}\plantag{[STEP\_END]} \plantag{[STEP\_BEGIN]} {\bfseries\color{diffcol!85!black}sg1.1.step2}\\
\hspace*{2.4em}\plantag{[ACTION]} \textbf{craft} \plantag{[/ACTION]} \plantag{[ARGS]} \planargtext{2 pink dye using}\\
\hspace*{3em}\planargtext{1 pink tulip} \plantag{[/ARGS]}\\
\hspace*{2.4em}\plantag{[EXPECTED\_OBS]} \planobstext{Crafted 2 minecraft:pink\_dye}\\
\hspace*{3em}\plantag{[/EXPECTED\_OBS]} \plantag{[STEP\_END]} \plantag{[SG\_END]}\\
\hspace*{1.2em}\plantag{[SG\_BEGIN op=SEQ]} {\bfseries\color{diffcol!85!black}sg1.2}\\
\hspace*{1.8em}\plantag{[SG\_INTENT]} \plansubgoaltext{fetch sand and gravel} \plantag{[/SG\_INTENT]}\\
\hspace*{1.8em}\plantag{[STEP\_BEGIN]} {\bfseries\color{diffcol!85!black}sg1.2.step1}\\
\hspace*{2.4em}\plantag{[ACTION]} \textbf{fetch} \plantag{[/ACTION]} \plantag{[ARGS]} \planargtext{4 sand} \plantag{[/ARGS]}\\
\hspace*{2.4em}\plantag{[EXPECTED\_OBS]} \planobstext{Got 4 sand} \plantag{[/EXPECTED\_OBS]}\\
\hspace*{1.8em}\plantag{[STEP\_END]} \plantag{[STEP\_BEGIN]} {\bfseries\color{diffcol!85!black}sg1.2.step2}\\
\hspace*{2.4em}\plantag{[ACTION]} \textbf{fetch} \plantag{[/ACTION]} \plantag{[ARGS]} \planargtext{4 gravel} \plantag{[/ARGS]}\\
\hspace*{2.4em}\plantag{[EXPECTED\_OBS]} \planobstext{Got 4 gravel} \plantag{[/EXPECTED\_OBS]}\\
\hspace*{2.4em}\plantag{[STEP\_END]} \plantag{[SG\_END]} \plantag{[SG\_END]}\\
\hspace*{0.6em}\plantag{[SG\_BEGIN op=SEQ]} {\bfseries\color{diffcol!85!black}sg2}\\
\hspace*{1.2em}\plantag{[SG\_INTENT]} \plansubgoaltext{craft pink concrete powder from pink dye,}\\
\hspace*{1.8em}\plansubgoaltext{sand, and gravel} \plantag{[/SG\_INTENT]}\\
\hspace*{1.2em}\plantag{[STEP\_BEGIN]} {\bfseries\color{diffcol!85!black}sg2.step1}\\
\hspace*{1.8em}\plantag{[ACTION]} \textbf{craft} \plantag{[/ACTION]} \plantag{[ARGS]} \planargtext{8 pink concrete powder}\\
\hspace*{2.4em}\planargtext{using 1 pink dye, 4 sand, 4 gravel} \plantag{[/ARGS]}\\
\hspace*{1.8em}\plantag{[EXPECTED\_OBS]} \planobstext{Crafted 8 minecraft:pink\_concrete\_powder}\\
\hspace*{2.4em}\plantag{[/EXPECTED\_OBS]} \plantag{[STEP\_END]} \plantag{[SG\_END]}\\
\strut\plantag{[PLAN\_END]}};
\begin{scope}[shift={($(0,0 |- listing.center)+(9.50,2.45)$)},
 treebox/.style={draw=rulecol,solid,line width=.55pt,rounded corners=1.2pt,
   align=center,inner sep=3pt,font=\sffamily\fontsize{8}{9.5}\selectfont,text=ink},
 sg/.style={treebox,fill=diffcol!6},
 leaf/.style={treebox,fill=white},
 edge/.style={-{Stealth[length=1.2mm,width=.85mm]},draw=muted,line width=.55pt,
   shorten <=1pt,shorten >=1pt}]
\node[fp title] at (0,1.35) {(b) The same plan, parsed};
\node[sg,fill=plangoalcol!6,text width=3.7cm] (goal) at (4.05,0)
 {\textcolor{plangoalcol}{\textbf{Goal}}\\craft pink concrete powder};
\node[sg,text width=2.45cm] (sg1) at (2.95,-1.40)
 {\textbf{sg1}\\Prepare all materials};
\node[sg,text width=1.70cm] (sg2) at (6.80,-1.40)
 {\textbf{sg2}\\Craft final item};
\node[sg,text width=1.75cm] (sg11) at (1.15,-2.90)
 {\textbf{sg1.1}\\Prepare pink dye};
\node[sg,text width=1.75cm] (sg12) at (4.75,-2.90)
 {\textbf{sg1.2}\\Fetch sand and gravel};
\node[leaf,text width=1.70cm,anchor=north] (final) at (6.80,-2.52)
 {\textcolor{diffcol}{\texttt{step1}}\\\textbf{craft}\\8 pink concrete powder};
\node[leaf,text width=1.35cm,anchor=north] (l11) at (.25,-4.45)
 {\textcolor{diffcol}{\texttt{step1}}\\\textbf{fetch}\\1 pink tulip};
\node[leaf,text width=1.35cm,anchor=north] (l12) at (2.05,-4.45)
 {\textcolor{diffcol}{\texttt{step2}}\\\textbf{craft}\\2 pink dye};
\node[leaf,text width=1.35cm,anchor=north] (l21) at (3.85,-4.45)
 {\textcolor{diffcol}{\texttt{step1}}\\\textbf{fetch}\\4 sand};
\node[leaf,text width=1.35cm,anchor=north] (l22) at (5.65,-4.45)
 {\textcolor{diffcol}{\texttt{step2}}\\\textbf{fetch}\\4 gravel};
\foreach \parent/\child in {goal/sg1,goal/sg2,sg1/sg11,sg1/sg12,sg11/l11,sg11/l12,sg12/l21,sg12/l22,sg2/final}
 \draw[edge] (\parent.south) -- (\child.north);
\node[anchor=north west,align=left,text width=7.15cm,inner sep=0pt,
 font=\sffamily\fontsize{7.5}{9}\selectfont,text=muted] at (-.35,-6.05)
 {Arrows show containment; subgoals are shaded.};
\end{scope}
\end{tikzpicture}

%% file: method.tex
\section{The \sys Framework}
\label{sec:method}
\suppressfloats[t]

\sys follows a plan-and-act architecture \citep{adaplanner,planandact}, with
one diffusion planner jointly trained for plan generation and repair
(\Cref{fig:framework}). The planner generates a complete structured plan,
which a separate executor translates into environment actions. After failure,
the executor selects a repair region and supplies repair feedback. The planner
infills this region while preserving the surrounding plan text.

\subsection{Plan generation and repair}
\label{sec:operations}

\paragraph{Structured plan representation.}
\label{sec:planrep}
The hierarchy in \Cref{def:plan,fig:plantree} provides explicit boundaries
for selecting and replacing plan text. Step and subgoal identifiers locate
spans for masking during training or repair. A repair region may include a
step or an enclosing subgoal, together with associated reasoning and intent
fields. This supports repair at the step or subgoal level while preserving the
surrounding text.

\paragraph{Task 1: Plan generation.}
\label{sec:generate}
Given a task goal, the initial environment observation, and the available
actions, the planner generates a complete plan using the generation procedure
in \Cref{sec:decoding}. The executor follows the plan strictly, carrying out
its leaf steps in order. The diffusion and AR planners receive these same inputs
and use the same structured plan representation.

\paragraph{Failure detection.}
An execution fails if it terminates without achieving the task goal.
The \emph{execution history} $\exechistory$ records ordered actions and their
\emph{environment feedback}, comprising observations and error messages returned
by the environment. We separately prompt the executor to select steps in the
failed plan $\plan$ for revision using $\exechistory$, or to abstain.
Selected steps and associated fields define a repair region
$\repairregion\subseteq[\planlength]$, where $[\planlength]=\{1,\dots,\planlength\}$.
The detector supplies \emph{repair feedback} $\repairfeedback$, combining
environment feedback with an explanation of why the selected text needs
revision. Failure may be observed at a different step from the one requiring
repair.

This procedure requires no reference plan (\Cref{app:detector}). For training
and a separate reference-assisted evaluation, a language-model judge uses a
reference plan to select repair regions and supply repair feedback. In that
evaluation, both planners receive the same failed plans, selected regions,
and repair feedback.

\begin{samepage}
\paragraph{Task 2: Plan repair.}
\label{sec:repair}
The planner receives the failed plan $\plan$, selected region $\repairregion$,
and repair feedback $\repairfeedback$. For a contiguous region, write
$\plan=\plan_{\mathrm{pre}}\Vert\plan_{\repairregion}\Vert\plan_{\mathrm{post}}$,
where $\Vert$ denotes concatenation. Conditioned on the preserved prefix,
suffix, and repair feedback, the planner generates a replacement $\replacement$ to form
\begin{equation}
\label{eq:localized-repair}
\plan'=\plan_{\mathrm{pre}}\Vert\replacement\Vert\plan_{\mathrm{post}}.
\end{equation}
\end{samepage}
The replacement may differ in length from the selected span. For multiple
spans, both planners jointly generate replacements, preserving their order and all intervening text;
$\repairregion$ indexes tokens in the original plan.

The diffusion planner allocates a total of $\repaircapacity=2|\repairregion|$
replacement positions across the selected spans, where $|\repairregion|$ counts tokens in
the original repair region. The planner then applies the infilling procedure
in \Cref{sec:decoding}. Each span has its own padded canvas segment, and its
replacement ends before the first \texttt{[delete]} token in that segment.
The diffusion and AR planners receive the same preserved plan context and
repair feedback. In our interactive evaluation, each repaired plan is executed from
the task's initial environment state (\Cref{app:envs}).
\Cref{fig:teaser} shows an ALFWorld example.

\subsection{Reference plans and repair examples}
\label{sec:training-data}
\label{sec:setup-harvest}

We partition each environment's tasks into disjoint \emph{train-gen},
\emph{train-rep}, \emph{dev}, and \emph{test} sets. The first two supply
complete-plan targets and repair examples, respectively; \emph{dev} determines
training convergence and \emph{test} is reserved for evaluation.

\paragraph{Reference-plan construction.}
We replay successful action sequences, storing each action, its arguments, and
observed feedback as a leaf step (\Cref{fig:reference-plans}). Fixed domain rules
group leaves into subgoals. An LLM rewrites only reasoning and intents; invalid
rewrites revert to the original plan. A separate executor checks for positive
terminal reward. Accepted \emph{train-gen} plans provide generation targets;
\emph{train-rep} plans provide repair references (\Cref{app:envs}).

\input{figs/fig_reference_plans.tex}

\paragraph{Repair examples.}
We train generation-only DreamReasoner-8B and Qwen3-8B planners per environment
on \emph{train-gen} and execute them on \emph{train-rep}, retaining failed plans
and their execution histories.
A reference-assisted judge selects defective regions and supplies repair
feedback; corresponding reference text provides the replacement targets.
We supplement these real failures with structured perturbations of verified
training plans, changing an object, receptacle, intent, or prerequisite.
The original text in the perturbed region supplies the replacement target.
We reject examples with edits outside the selected region or replacements
appearing verbatim in the visible context. Both model families receive the
same generation and repair examples.

\subsection{Joint training and decoding}
\label{sec:training}

\paragraph{Retaining both capabilities.}
Each final planner starts from its original DreamReasoner-8B or Qwen3-8B
checkpoint and is trained jointly on generation and repair for its
environment (\Cref{fig:curriculum}). In preliminary experiments with both diffusion and AR planners,
sequentially fine-tuning a generation-only planner on repair examples left repair
performance roughly unchanged but substantially degraded complete-plan
generation, primarily through invalid formatting. This finding motivated
joint training on the combined generation and repair examples.

\paragraph{Targets and block sizes.}
Generation conditions on the goal, initial observation, and action grammar;
its target is the complete plan. Repair conditions on the failed plan's
preserved context and repair feedback; only replacement targets contribute
to the repair loss. For diffusion, target tokens are corrupted and supervised
using the masked-token objective in \Cref{sec:formal}. DreamReasoner trains
with progressively larger blocks \citep{dreamreasoner}. We adapt this block-size
curriculum with separate sampling distributions for generation and repair
(\Cref{fig:curriculum}). Generation assigns
$61.2\%$ of its sampling weight to blocks of at most $16$ tokens, whereas
repair assigns $49.0\%$ to the $4{,}096$-token block. This emphasizes short
blocks for constructing plans and larger blocks for infilling supplied
regions with fixed surrounding context. The AR planner learns plan generation
and repair on the same examples, using FIM serialization for repair.

\input{figs/fig_curriculum.tex}

\paragraph{Inference.}
On ALFWorld and TextCraft, diffusion generation uses left-to-right blocks of
$4$ tokens with confidence threshold $0.95$. Repair uses full-canvas infilling
with RelaxedConfidence, threshold floor and neighbour-support cutoff $0.7$,
and neighbourhood radius $4$ (\Cref{sec:decoding,tab:hparams}). Small generation
blocks limit how much of the plan is resolved together; repair uses the
preserved prefix and suffix throughout infilling.
The AR planner jointly generates replacements for the selected spans in one left-to-right
sequence. \Cref{app:hparams} specifies optimization, padding, and prompts.

%% file: figs/fig_reference_plans.tex
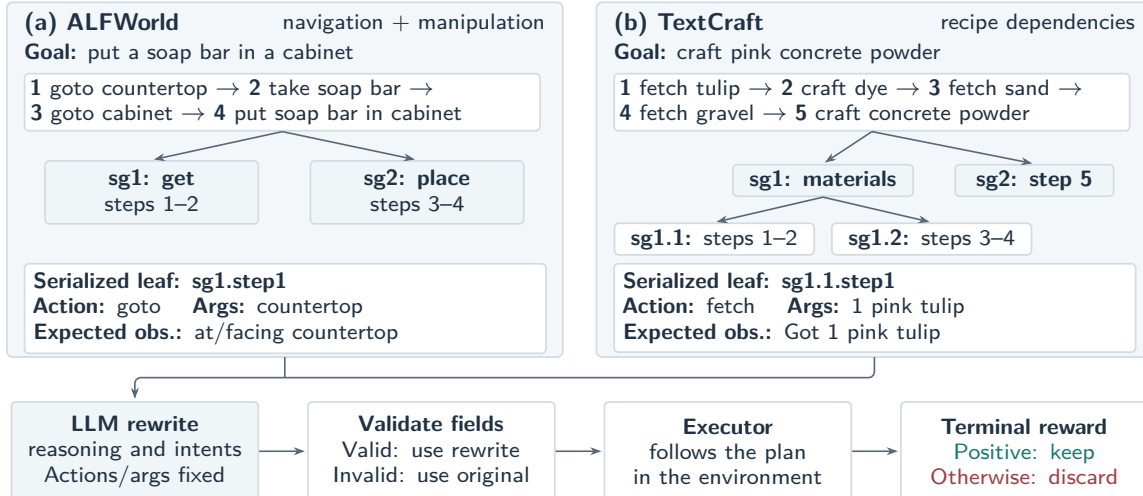
\begin{figure}[t]
\centering
\resizebox{0.94\linewidth}{!}{\input{figure_assets/reference_plans.tikz}}
\caption{\textbf{Reference-plan construction.} ALFWorld groups navigation--manipulation
pairs; TextCraft groups ingredient preparation before the final craft.
Numbers map actions to subgoals; leaf boxes show serialized fields.
Both domains share the bottom pipeline. Actions and observations are abbreviated.}
\label{fig:reference-plans}
\end{figure}

%% file: figure_assets/reference_plans.tikz
\begin{tikzpicture}[x=1cm,y=.94cm,
 rp text/.style={font=\sffamily\fontsize{7.5}{8.6}\selectfont,text=ink},
 rp title/.style={rp text,font=\sffamily\bfseries\fontsize{8.2}{9.4}\selectfont},
 rp box/.style={rp text,draw=rulecol,fill=white,rounded corners=1.5pt,
 line width=.6pt,align=center,inner sep=3pt},
 rp arrow/.style={-{Stealth[length=1.35mm,width=.95mm]},draw=muted,line width=.55pt,
 shorten >=.7pt},
 rp edge/.style={draw=muted,line width=.55pt},
 rp small/.style={rp text,font=\sffamily\fontsize{7.5}{8.6}\selectfont}]
\path[use as bounding box] (0,0) rectangle (13.8,6.55);
\foreach \x in {.10,7.10}
 \draw[draw=rulecol,fill=panelbg,rounded corners=2pt,line width=.6pt]
 (\x,1.99) rectangle +(6.60,4.48);
\node[rp title,anchor=west,inner sep=0pt] at (.30,6.20) {(a) ALFWorld};
\node[rp title,anchor=west,inner sep=0pt] at (7.30,6.20) {(b) TextCraft};
\node[rp text,anchor=west,inner sep=0pt] at (.30,5.82) {\textbf{Goal:} put a soap bar in a cabinet};
\node[rp text,anchor=west,inner sep=0pt] at (7.30,5.82) {\textbf{Goal:} craft pink concrete powder};
\node[rp box,rp small,text width=5.98cm,align=left,anchor=north west,inner sep=2pt] (aseq) at (.30,5.58)
 {\textbf{1} goto countertop {\fontsize{8}{9}\selectfont$\rightarrow$} \textbf{2} take soap bar {\fontsize{8}{9}\selectfont$\rightarrow$}\\
  \textbf{3} goto cabinet {\fontsize{8}{9}\selectfont$\rightarrow$} \textbf{4} put soap bar in cabinet};
\node[rp box,rp small,text width=5.98cm,align=left,anchor=north west,inner sep=2pt] (tseq) at (7.30,5.58)
 {\textbf{1} fetch tulip {\fontsize{8}{9}\selectfont$\rightarrow$} \textbf{2} craft dye {\fontsize{8}{9}\selectfont$\rightarrow$} \textbf{3} fetch sand {\fontsize{8}{9}\selectfont$\rightarrow$}\\
  \textbf{4} fetch gravel {\fontsize{8}{9}\selectfont$\rightarrow$} \textbf{5} craft concrete powder};
\node[rp small,anchor=east,inner sep=0pt] at (6.50,6.20) {navigation + manipulation};
\node[rp small,anchor=east,inner sep=0pt] at (13.50,6.20) {recipe dependencies};

\node[rp box,rp small,fill=diffcol!7,text width=2.32cm] (apick) at (1.82,4.07)
 {\textbf{sg1: get}\\steps 1--2};
\node[rp box,rp small,fill=diffcol!7,text width=2.32cm] (aput) at (4.97,4.07)
 {\textbf{sg2: place}\\steps 3--4};
\draw[rp arrow] (aseq.south) -- (apick.north);
\draw[rp arrow] (aseq.south) -- (aput.north);

\node[rp box,rp small,fill=diffcol!7,text width=2.00cm,inner sep=2pt] (tmat) at (9.79,4.22)
 {\textbf{sg1: materials}};
\node[rp box,rp small,fill=diffcol!7,text width=1.74cm,inner sep=2pt] (tfinal) at (12.30,4.22)
 {\textbf{sg2: step 5}};
\node[rp box,rp small,text width=2.23cm,inner sep=2pt] (tdye) at (8.50,3.48)
 {\textbf{sg1.1:} steps 1--2};
\node[rp box,rp small,text width=2.23cm,inner sep=2pt] (tsand) at (11.08,3.48)
 {\textbf{sg1.2:} steps 3--4};
\draw[rp arrow] (tseq.south) -- (tmat.north);
\draw[rp arrow] (tseq.south) -- (tfinal.north);
\draw[rp arrow] (tmat.south) -- (tdye.north);
\draw[rp arrow] (tmat.south) -- (tsand.north);

\node[rp box,rp small,anchor=north west,align=left,text width=5.98cm] at (.30,3.18)
 {\textbf{Serialized leaf: sg1.step1}\\
  \textbf{Action:} goto \quad \textbf{Args:} countertop\\
  \textbf{Expected obs.:} at/facing countertop};
\node[rp box,rp small,anchor=north west,align=left,text width=5.98cm] at (7.30,3.18)
 {\textbf{Serialized leaf: sg1.1.step1}\\
  \textbf{Action:} fetch \quad \textbf{Args:} 1 pink tulip\\
  \textbf{Expected obs.:} Got 1 pink tulip};

\node[rp box,rp small,fill=diffcol!7,text width=2.72cm,minimum height=1.17cm] (rewrite) at (1.62,.80)
 {\textbf{LLM rewrite}\\reasoning and intents\\Actions/args fixed};
\node[rp box,rp small,text width=2.72cm,minimum height=1.17cm] (validate) at (5.14,.80)
 {\textbf{Validate fields}\\Valid: use rewrite\\Invalid: use original};
\node[rp box,rp small,text width=2.72cm,minimum height=1.17cm] (execute) at (8.66,.80)
 {\textbf{Executor}\\follows the plan\\in the environment};
\node[rp box,rp small,text width=2.72cm,minimum height=1.17cm] (filter) at (12.18,.80)
 {\textbf{Terminal reward}\\\textcolor{repaircol}{Positive: keep}\\\textcolor{errorcol}{Otherwise: discard}};
\draw[rp edge,rounded corners=2pt]
 (3.40,1.99) -- (3.40,1.73);
\draw[rp edge,rounded corners=2pt]
 (10.40,1.99) -- (10.40,1.73) -- (3.40,1.73);
\draw[rp arrow,rounded corners=2pt]
 (3.40,1.73) -- (1.62,1.73) -- (rewrite.north);
\foreach \a/\b in {rewrite/validate,validate/execute,execute/filter}
 \draw[rp arrow] (\a.east) -- (\b.west);
\end{tikzpicture}

%% file: figs/fig_curriculum.tex
\begin{figure}[t]
\centering
\resizebox{\linewidth}{!}{\input{figure_assets/curriculum.tikz}}
\caption{\textbf{Joint training with task-specific block sizes.}
The initial planner supplies failures for data construction; the final planner
starts from the original checkpoint and learns generation and repair jointly.
We adapt DreamReasoner's block-size curriculum \citep{dreamreasoner} with
separate sampling distributions for generation and repair. The lower panel
shows weights normalized within each objective; dashes denote unsampled sizes.
Inference uses the separate settings in \Cref{tab:hparams}.}
\label{fig:curriculum}
\end{figure}

%% file: figure_assets/curriculum.tikz
\begin{tikzpicture}[x=1cm,y=1cm,
  ct/.style={font=\sffamily\fontsize{8}{9.5}\selectfont,text=ink},
  title/.style={ct,font=\sffamily\bfseries\fontsize{8.4}{10}\selectfont},
  arr/.style={-{Stealth[length=1.6mm]},draw=muted,line width=.65pt}]
\path[use as bounding box] (0,0) rectangle (13.8,4.65);
\draw[draw=rulecol,fill=panelbg,rounded corners=2pt] (.05,3.17) rectangle (5.92,4.53);
\draw[draw=rulecol,fill=panelbg,rounded corners=2pt] (7.65,3.17) rectangle (13.75,4.53);
\node[title] at (2.99,4.18) {1. Collect training failures};
\node[ct,align=center] at (2.99,3.65)
  {Train an initial generation-only planner;\\execute it on disjoint repair-training tasks.};
\node[title] at (10.70,4.18) {2. Train the final planner jointly};
\node[ct,align=center] at (10.70,3.65)
  {Initialize from the original checkpoint;\\jointly train generation and repair.};
\draw[arr] (5.92,3.66) -- (7.65,3.66);
\node[ct,align=center,font=\sffamily\fontsize{7}{8}\selectfont] at (6.79,4.15) {Repair\\examples};
\node[ct,anchor=west,inner sep=0pt] at (.05,2.72)
  {\textbf{Diffusion block-size sampling within each objective} (normalized weights, \%)};
\node[ct,anchor=east] at (2.06,2.17) {Block size};
\foreach \size [count=\i from 0] in {4,8,16,32,64,128,256,512,1024,2048,4096}{
  \node[ct] at ({2.65+.99*\i},2.17) {\size};
}
\node[title,anchor=east] at (2.06,1.57) {Generation};
\node[title,anchor=east] at (2.06,.97) {Repair};
\foreach \val/\shade [count=\i from 0] in {24.5/49,20.4/41,16.3/33,12.2/24,8.2/16,6.1/12,5.1/10,4.1/8,3.1/6,--/0,--/0}{
  \fill[diffcol!\shade!white] ({2.18+.99*\i},1.32) rectangle ({3.12+.99*\i},1.82);
  \node[ct] at ({2.65+.99*\i},1.57) {\val};
}
\foreach \val/\shade [count=\i from 0] in {--/0,--/0,--/0,4.1/8,6.1/12,8.2/16,12.2/24,--/0,--/0,20.4/41,49.0/75}{
  \fill[diffcol!\shade!white] ({2.18+.99*\i},.72) rectangle ({3.12+.99*\i},1.22);
  \node[ct] at ({2.65+.99*\i},.97) {\val};
}
\node[ct,anchor=west,inner sep=0pt] at (.05,.27)
  {Generation supervises the complete plan; repair supervises only replacement spans.};
\end{tikzpicture}

%% file: experiments.tex
\section{Experiments}
\label{sec:exp}

We evaluate whether the jointly trained planners generate successful plans
at lower latency, how repair depends on region selection, and how performance
changes with the executor and plan representation. Natural Plan provides a
separate evaluation of generation and repair without task-specific training.

\subsection{Experimental settings}
\label{sec:setup}

\paragraph{Benchmarks and data.}
ALFWorld \citep{alfworld} requires household navigation and manipulation;
TextCraft \citep{adapt} requires dependency-constrained crafting.
Task success means that executing the plan achieves the environment's goal.
\Cref{tab:datasets} reports task splits and repair corpora constructed by
\Cref{sec:training-data}. Task counts precede reference-plan execution filtering
(\Cref{fig:reference-plans});
\Cref{tab:goldquality} reports candidate acceptance rates.

The \emph{train-gen} and \emph{train-rep} partitions contain $2{,}665/888$ ALFWorld
and $3{,}824/1{,}268$ TextCraft tasks, respectively. Repair corpora contain
$699$ real plus $763$ synthetic examples for ALFWorld, and $267$ real plus
$1{,}661$ synthetic examples for TextCraft.

\paragraph{Models and optimization.}
The diffusion planner \diffp is DreamReasoner-8B \citep{dreamreasoner}, adapted
from Qwen3-8B-Base. The AR planner
\arp is Qwen3-8B. Both have $8$B parameters and share a tokenizer, task
splits, training examples, and plan representation. We fully fine-tune each
planner per environment with AdamW until development-set performance converges.
Their prior training, decoding, and training compute differ.
\Cref{sec:training,tab:hparams} give block sizes and optimization settings.
A shared Qwen3-8B executor receives no task-specific fine-tuning. ReAct
\citep{react} uses the same Qwen3-8B model to act without a separate plan.

\paragraph{Evaluation protocol.}
Generation uses $134$ ALFWorld and $100$ TextCraft test tasks, with limits of
$60$ planned leaf steps and $1{,}000$ episode tokens. For repair, we pool
failed plans from both planners: $148$ ALFWorld and $71$ TextCraft failures.
Each example is identified by task and generating planner, so one task may contribute
two failed plans. Reference-assisted region selection uses all $219$ failures.
History-based selection uses the $175$ failures with step-level reference
annotations ($126$ ALFWorld, $49$ TextCraft), which support evaluation of the
selected regions; the detector itself sees only the plan and execution history.
Both planners receive the same failed plans, selected regions, and
repair feedback. Each example permits one repair attempt followed by execution from
the initial state. Missing regions and abstentions count as unsuccessful.
\Cref{app:envs} gives the complete population definitions.

\subsection{Plan quality and latency}
\label{sec:exp-main}

\input{figs/tab_main.tex}

\paragraph{Joint training retains generation and repair.}
All $234$ generated plans from each final planner pass the parser's format
checks. Both planners outperform ReAct in generation, with a slight edge
to AR. The planners differ by at most $3$ percentage points in generation
and reference-assisted repair (\Cref{tab:main}). These results show that the joint
training recipe retains both generation and repair performance in a single checkpoint (\Cref{fig:curriculum}).
\Cref{app:results} reports paired generation outcomes and confidence intervals.

\paragraph{Parallel decoding reduces generation latency.}
Diffusion reduces mean generation latency by \mbox{$39$--$46\%$} across the two environments
(\Cref{fig:cost}). Latency is measured using synchronized, unbatched bf16 decoding
on one A100. Repair is faster than plan generation for both planners.
Diffusion therefore generates plans faster at similar observed task success.
\Cref{app:decode} reports replacement tokens committed per forward pass.

\input{figs/fig_cost.tex}

\paragraph{Repair without task-specific training.}
Natural Plan \citep{naturalplan} evaluates travel and scheduling plans using
programmatic constraint validators, without an executor. Repair examples blank
regions of reference plans; both planners receive the same context, repair
feedback, and approximate replacement length. We use the original checkpoints;
diffusion uses block size $4$ for generation and $256$ for repair.

AR has higher generation success, while diffusion nearly doubles repair
success; both judges agree with these rankings (\Cref{tab:np}).
Diffusion's repair advantage is concentrated in trip planning. Its flight-graph
structure favors flexible decoding, analogous to diffusion's reverse
decoding on synthetic Star-Path tasks \citep{asymmetry}.
\Cref{app:np} gives domain results and uncertainty estimates.

\input{figs/tab_np.tex}

\subsection{What limits execution and repair?}
\label{sec:diagnostics}

\paragraph{Region selection.}
On the same $175$ eligible failures, with repair feedback fixed, history-based
selection reduces pooled repair success relative to reference-assisted selection by $12$ percentage
points for diffusion and $8.6$ for AR
(\Cref{tab:detector}).
These results show that selecting the repair region correctly is important for successful
plan repair.

\paragraph{Alternative structured representation.}
We compare two structured formats: delimited plans and prose plans with phase
headings and numbered steps. We retrain both planners, keeping task goals,
executable fields, splits, and decoding settings fixed.
Under Qwen3-8B, prose plans improve generation success by $17.9$--$22.4$
percentage points on ALFWorld and $6$--$12$ on TextCraft
(\Cref{tab:nlgen}). Diffusion's observed reference-assisted repair
success is lower for prose than delimited plans: $21.3\%$ versus $28.4\%$
on ALFWorld and $5.0\%$ versus $15.5\%$ on TextCraft, evaluated on their
respective failure sets.
AR has higher prose-plan repair success than diffusion in both environments.
Plan format thus affects generation and repair differently: explicit boundaries
may simplify repair-region selection and support recovery.
\Cref{app:nlsurface} gives complete results and executor comparisons.

\paragraph{Error analysis.}
Despite AR's slight aggregate lead, its successes do not subsume diffusion's:
diffusion solves $23/134$ ALFWorld and $13/100$ TextCraft tasks that AR fails
(\Cref{tab:paired-generation}). Cooling favors AR and cleaning favors diffusion.
Among tasks failed by only one planner, failure at a leaf step is the most
common stopping reason for both; invalid-action limits are also frequent in
ALFWorld (\Cref{app:errors}).

\begin{table}[H]
\centering
\caption{\textbf{Pairwise error analysis of generated plans.} Counts partition
the test tasks. In each ``only'' column, the other planner succeeds.}
\label{tab:paired-generation}
\small
\setlength{\tabcolsep}{5pt}
\begin{tabular}{lrrrr}
\toprule
Environment (tasks) & Both succeed & Both fail & Only diffusion fails & Only AR fails \\
\midrule
ALFWorld ($134$) & $36$ & $50$ & $25$ & $23$ \\
TextCraft ($100$) & $50$ & $21$ & $16$ & $13$ \\
\bottomrule
\end{tabular}
\end{table}

%% file: figs/tab_main.tex
\begin{table}[!htbp]
\centering
\caption{\textbf{Task success under the Qwen3-8B executor (\%).}
Generation: $134/100$ tasks; reference-assisted/history-based repair: $148/71$ and
$126/49$ failures (ALFWorld/TextCraft), with one attempt and restart.
\textsuperscript{$\dagger$}End-to-end provides generation-plus-repair success
including recovery on each planner's own failures using the history-based repair. Dashes denote no repair stage.}
\label{tab:main}
\label{tab:appcounts}
\small
\setlength{\tabcolsep}{5pt}
\begin{tabular}{llcccc}
\toprule
Environment & Model & \shortstack{Plan\\generation} & \shortstack{Reference-assisted\\repair} & \shortstack{History-based\\repair} & End-to-end\textsuperscript{$\dagger$} \\
\midrule
\multirow{3}{*}{ALFWorld}
 & ReAct  & $38.1\%$ & $-$ & $-$ & $-$ \\
 & \diffp & $44.0\%$ & $28.4\%$ & $14.3\%$ & $59.9\%$ \\
 & \arp   & $45.5\%$ & $27.7\%$ & $17.5\%$ & $60.6\%$ \\
\midrule
\multirow{3}{*}{TextCraft}
 & ReAct  & $48.0\%$ & $-$ & $-$ & $-$ \\
 & \diffp & $63.0\%$ & $15.5\%$ & $24.5\%$ & $68.7\%$ \\
 & \arp   & $66.0\%$ & $18.3\%$ & $30.6\%$ & $72.2\%$ \\
\bottomrule
\end{tabular}
\end{table}

%% file: figs/fig_cost.tex
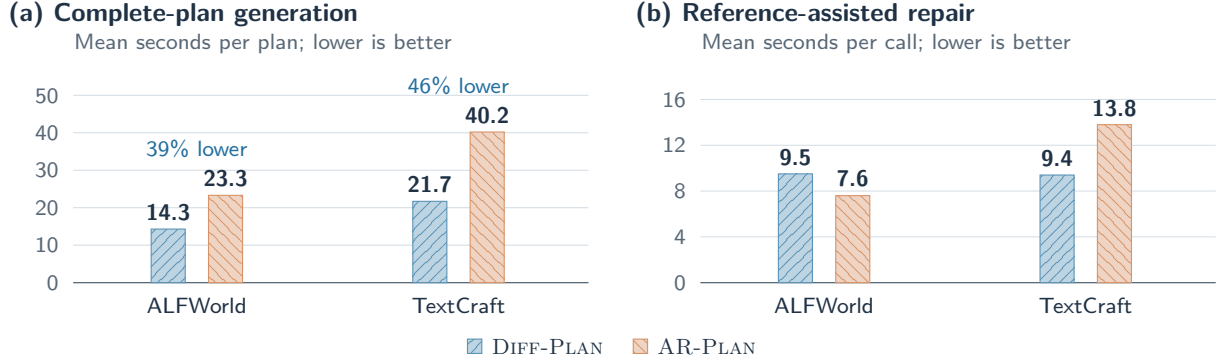
\begin{figure}[t]
\centering
\resizebox{\linewidth}{!}{\input{figure_assets/cost.tikz}}
\caption{\textbf{Diffusion reduces mean plan-generation latency by $39$--$46\%$.} Values are synchronized, unbatched calls on one A100 in bf16; prompt construction, parsing, and environment execution are excluded. Repair latency varies by environment.}
\label{fig:cost}
\end{figure}

%% file: figure_assets/cost.tikz
\begin{tikzpicture}[x=1cm,y=1cm,fp text]
\path (0,-0.95) rectangle (13.8,3.13);
\begin{scope}[xshift=0cm]
\node[fp title] at (0.02,2.99) {(a) Complete-plan generation};
\node[fp label,anchor=west] at (0.72,2.67) {Mean seconds per plan; lower is better};
\draw[fp grid] (0.75,0.0000)--(6.50,0.0000);
\node[fp label,anchor=east] at (0.63,0.0000) {0};
\draw[fp grid] (0.75,0.4182)--(6.50,0.4182);
\node[fp label,anchor=east] at (0.63,0.4182) {10};
\draw[fp grid] (0.75,0.8364)--(6.50,0.8364);
\node[fp label,anchor=east] at (0.63,0.8364) {20};
\draw[fp grid] (0.75,1.2545)--(6.50,1.2545);
\node[fp label,anchor=east] at (0.63,1.2545) {30};
\draw[fp grid] (0.75,1.6727)--(6.50,1.6727);
\node[fp label,anchor=east] at (0.63,1.6727) {40};
\draw[fp grid] (0.75,2.0909)--(6.50,2.0909);
\node[fp label,anchor=east] at (0.63,2.0909) {50};
\draw[fp axis] (0.75,0)--(6.50,0);
\node[fp label,anchor=north,text=ink] at (2.13,-0.13) {ALFWorld};
\path[fp diff bar] (1.6200,0) rectangle (2.0000,0.5980);
\node[fp value,anchor=south] at (1.8100,0.6580) {14.3};
\path[fp ar bar] (2.2600,0) rectangle (2.6400,0.9744);
\node[fp value,anchor=south] at (2.4500,1.0344) {23.3};
\node[fp label,text=diffcol,anchor=south] at (2.13,1.3544) {39\% lower};
\node[fp label,anchor=north,text=ink] at (5.05,-0.13) {TextCraft};
\path[fp diff bar] (4.5400,0) rectangle (4.9200,0.9075);
\node[fp value,anchor=south] at (4.7300,0.9675) {21.7};
\path[fp ar bar] (5.1800,0) rectangle (5.5600,1.6811);
\node[fp value,anchor=south] at (5.3700,1.7411) {40.2};
\node[fp label,text=diffcol,anchor=south] at (5.05,2.0611) {46\% lower};
\end{scope}
\begin{scope}[xshift=7.0cm]
\node[fp title] at (0.02,2.99) {(b) Reference-assisted repair};
\node[fp label,anchor=west] at (0.72,2.67) {Mean seconds per call; lower is better};
\draw[fp grid] (0.75,0.0000)--(6.50,0.0000);
\node[fp label,anchor=east] at (0.63,0.0000) {0};
\draw[fp grid] (0.75,0.5111)--(6.50,0.5111);
\node[fp label,anchor=east] at (0.63,0.5111) {4};
\draw[fp grid] (0.75,1.0222)--(6.50,1.0222);
\node[fp label,anchor=east] at (0.63,1.0222) {8};
\draw[fp grid] (0.75,1.5333)--(6.50,1.5333);
\node[fp label,anchor=east] at (0.63,1.5333) {12};
\draw[fp grid] (0.75,2.0444)--(6.50,2.0444);
\node[fp label,anchor=east] at (0.63,2.0444) {16};
\draw[fp axis] (0.75,0)--(6.50,0);
\node[fp label,anchor=north,text=ink] at (2.13,-0.13) {ALFWorld};
\path[fp diff bar] (1.6200,0) rectangle (2.0000,1.2139);
\node[fp value,anchor=south] at (1.8100,1.2739) {9.5};
\path[fp ar bar] (2.2600,0) rectangle (2.6400,0.9711);
\node[fp value,anchor=south] at (2.4500,1.0311) {7.6};
\node[fp label,anchor=north,text=ink] at (5.05,-0.13) {TextCraft};
\path[fp diff bar] (4.5400,0) rectangle (4.9200,1.2011);
\node[fp value,anchor=south] at (4.7300,1.2611) {9.4};
\path[fp ar bar] (5.1800,0) rectangle (5.5600,1.7633);
\node[fp value,anchor=south] at (5.3700,1.8233) {13.8};
\end{scope}
\path[fp diff bar] (5.15,-0.79) rectangle (5.31,-0.63);
\node[fp label,anchor=west,text=ink] at (5.38,-0.71) {\diffp};
\path[fp ar bar] (7.00,-0.79) rectangle (7.16,-0.63);
\node[fp label,anchor=west,text=ink] at (7.23,-0.71) {\arp};
\end{tikzpicture}

%% file: figs/tab_np.tex
\begin{table}[H]
\centering
\caption{\textbf{Natural Plan without task-specific training.} Acceptance
rates (\%) under the benchmark validator and two judges. Generation uses
$1{,}000$ tasks; repair infills reference-plan regions in $2{,}661$ examples
from $696$ tasks.}
\label{tab:np}
\small
\setlength{\tabcolsep}{8pt}
\begin{tabular}{llrrr}
\toprule
Task & Model & Validator & GPT-5 & Sonnet 5 \\
\midrule
\multirow{2}{*}{Generation}
 & \diffp & $33.0\%$ & $20.2\%$ & $31.8\%$ \\
 & \arp   & $48.6\%$ & $34.0\%$ & $46.1\%$ \\
\midrule
\multirow{2}{*}{Repair}
 & \diffp & $53.7\%$ & $72.9\%$ & $81.6\%$ \\
 & \arp   & $27.0\%$ & $42.1\%$ & $53.4\%$ \\
\bottomrule
\end{tabular}
\end{table}

%% file: conclusion.tex
\section{Conclusion and Future Work}
\label{sec:conclusion}

\sys combines complete-plan generation with repair through infilling,
preserving unselected text. On ALFWorld and TextCraft, the planners achieve
similar observed generation success, while diffusion reduces mean generation
latency by $39$--$46\%$. History-based repair enables recovery without reference
plans, with higher pooled success for AR. Without task-specific training on
Natural Plan, AR achieves higher generation success, whereas diffusion achieves higher repair
success. These findings support structured plans as an interface for parallel
generation and repair. Several directions remain for future work, including jointly learning
repair-region selection and infilling from execution feedback, adapting repair
scope and decoding effort to uncertainty, and extending the framework to longer
planning horizons while maintaining consistency between local edits and global
task goals.

%% file: appendix.tex
\section{Supplementary Material}
\raggedbottom

\renewcommand{\thefigure}{A.\arabic{figure}}
\renewcommand{\theHfigure}{A.\arabic{figure}}
\renewcommand{\theHtable}{A.\arabic{table}}
\renewcommand{\theequation}{A.\arabic{equation}}
\setcounter{figure}{0}

The appendix is organized as follows. \Cref{app:envs} defines the data and
evaluation populations. \Cref{app:results} reports the complete primary
results. \Cref{app:sensitivity} collects representation, executor,
region-selection, Natural Plan, and long-horizon analyses. \Cref{app:hparams}
records training, decoding, and prompt details.

\subsection{Protocols and data}
\label{app:envs}

ALFWorld combines household navigation and manipulation; TextCraft requires
dependency-constrained crafting. Natural Plan contains trip, meeting, and
calendar scheduling problems with programmatic validators that check whether
a completed plan satisfies the task constraints. ScienceWorld is
used only for the long-horizon generation analysis in \Cref{app:sw}.

\subsubsection{Task splits and training examples}

ALFWorld and TextCraft are partitioned by task identity into disjoint
\emph{train-gen}, \emph{train-rep}, \emph{dev}, and \emph{test} sets.
\emph{train-gen} supplies complete-plan targets. Generation-only DreamReasoner-8B
and Qwen3-8B checkpoints, trained separately per environment on that split,
produce failures on \emph{train-rep}. The \emph{dev} split determines training
convergence. Primary interactive evaluation uses \emph{test}.
Both model families use the same task splits and training examples.

\input{figs/tab_training_data.tex}

\paragraph{Dataset-specific reference construction.}
The benchmark task specifies the goal; the subgoal decomposition uses fixed,
domain-specific rules. ALFWorld references start from expert demonstrations. We group navigation
followed by manipulation into subgoals. TextCraft uses a deterministic solver
that expands recipe dependencies backward from the target item; these
dependencies define the subgoal hierarchy. In both environments, actions are
replayed to record observations. We render each action, its arguments, and its
observed environment feedback as a leaf step, then serialize these steps in execution
order. This constructs the initial plan deterministically from the source
sequence and replayed feedback.
Claude Sonnet 5 with default model settings rewrites reasoning and intent fields from the grouped plan
and its evidence while preserving actions and arguments. The returned fields
are rendered and validated; an invalid rewrite falls back to the
original structured plan. Claude Sonnet 5 with default model settings follows each
candidate as the verification executor in the live environment. Positive terminal reward is required for
acceptance (\Cref{fig:reference-plans}).

\begin{table}[H]
\centering
\caption{\textbf{Reference-plan candidates.} \emph{Train} is the union of
\emph{train-gen} and \emph{train-rep}. Counts precede execution filtering;
verified \emph{train-gen} candidates supply generation targets, while verified
\emph{train-rep} references provide repair targets for failed plans. Mean, median, and
maximum refer to the number of leaf steps; success is measured by execution.}
\label{tab:goldquality}
\small
\setlength{\tabcolsep}{6pt}
\begin{tabular}{llrrrrr}
\toprule
Environment & Split & Candidates & Success & Mean & Median & Max \\
\midrule
\multirow{4}{*}{ALFWorld}
 & train & $3{,}553$ & $91.0\%$ & $5.8$ & $6$ & $8$ \\
 & dev   & $140$     & $90.7\%$ & $5.7$ & $6$ & $8$ \\
 & test  & $134$     & $91.0\%$ & $5.6$ & $6$ & $8$ \\
 & all   & $3{,}827$ & $91.0\%$ & $5.8$ & $6$ & $8$ \\
\midrule
\multirow{4}{*}{TextCraft}
 & train & $5{,}092$ & $99.8\%$ & $9.8$ & $7.0$ & $66$ \\
 & dev   & $50$      & $100\%$  & $9.9$ & $7.0$ & $91$ \\
 & test  & $100$     & $99.0\%$ & $8.0$ & $6.5$ & $33$ \\
 & all   & $5{,}242$ & $99.8\%$ & $9.8$ & $7.0$ & $91$ \\
\bottomrule
\end{tabular}
\end{table}

For repair training, we retain failed plans with their execution history and
first observed failure. A reference-assisted judge identifies the smallest
addressable repair region associated with the failure. The corresponding text
from the verified reference plan provides the replacement target. Structured perturbations supplement these real failures.
They modify fields of the parsed plan, such as an object, receptacle, intent, or
prerequisite. An example is rejected if the edit changes text outside the selected
region or if its replacement occurs verbatim in the remaining prompt.

We query Claude Sonnet 5 (\texttt{us.anthropic.claude-sonnet-5}) once
with default model settings as the reference-assisted judge. It receives the goal, failed plan, execution history,
failure observation, and reference plan, and is instructed to return the
identifiers of the smallest set of steps whose differences from the reference
explain the observed failure, followed by a brief explanation.

\subsubsection{Interactive evaluation populations}

Generation is evaluated on $134$ ALFWorld and $100$ TextCraft test tasks.
The resulting failures are pooled across planners, giving $148$ failed plans
in ALFWorld ($75$ from \diffp and $73$ from \arp) and $71$ in TextCraft
($37$ from \diffp and $34$ from \arp). Each repair example is identified by
its task and generating planner, so a task failed by both planners contributes two
examples. All failed plans remain in the primary reference-assisted repair
population, including those with no usable repair region, which count as
unsuccessful.

History-based region selection is evaluated on $126/148$ ALFWorld and $49/71$
TextCraft failed plans with a step labelled by the reference-assisted judge.
These labels provide the step-level reference targets needed to evaluate region
selection. The other $44$ failed plans ($22$ from each environment) lack such a
label and are excluded from this comparison. On this subset of $175$ failed
plans, reference-assisted repair achieves task success with \diffp ($29.1\%$)
and \arp ($29.7\%$), compared with \diffp ($24.2\%$) and \arp ($24.7\%$) on
all $219$ primary failures. A stored failure-detector output is available for
every included example, and abstentions remain in the denominator.

Each failed plan permits one repair attempt. Both planners infill the selected
region, and the repaired plan is executed from the task's initial environment
state rather than resumed from the failure. The Qwen3-8B executor uses
temperature $0.6$ and top-$p$ $0.95$. Reported repair success is terminal task
success after this repair-and-restart procedure.

\subsection{Complete primary results}
\label{app:results}

\Cref{tab:main} reports success rates and evaluation denominators for generation
and both repair protocols. \Cref{fig:main} plots the generation and
reference-assisted repair results.

\input{figs/fig3_results.tex}

\begin{table}[H]
\centering
\caption{\textbf{Paired outcomes.} Generation compares the planners on the same test
tasks. Reference-assisted repair compares them on pooled failed plans, which
may include two plans from one task. Columns indicate which planners succeed.}
\label{tab:agreement}
\small
\setlength{\tabcolsep}{4.5pt}
\resizebox{\textwidth}{!}{%
\begin{tabular}{lrrrrrrrr}
\toprule
& \multicolumn{4}{c}{Plan generation} & \multicolumn{4}{c}{Reference-assisted repair} \\
\cmidrule(lr){2-5}\cmidrule(lr){6-9}
Environment & both & \diffp only & \arp only & neither & both & \diffp only & \arp only & neither \\
\midrule
ALFWorld  & $36$ & $23$ & $25$ & $50$ & $24$ & $18$ & $17$ & $89$ \\
TextCraft & $50$ & $13$ & $16$ & $21$ & $6$  & $5$  & $7$  & $53$ \\
\bottomrule
\end{tabular}}
\end{table}

A paired normal approximation based on tasks with different outcomes across
planners gives $95\%$ confidence intervals of $[-11.6,8.6]$ and $[-13.5,7.5]$ percentage points for the
diffusion-minus-AR generation difference on ALFWorld and TextCraft. These
intervals describe held-out task variation conditional on the trained checkpoints
and recorded execution outcomes. Repair examples are grouped by task, since a task may contribute a failed
plan from each planner.

Mean generation latency is lower for \diffp ($14.3$ seconds on ALFWorld and
$21.7$ on TextCraft) than for \arp ($23.3$ and $40.2$ seconds, respectively).
For reference-assisted repair, \arp ($7.6$ seconds) is faster than \diffp
($9.5$ seconds) on ALFWorld, while \diffp ($9.4$ seconds) is faster than \arp
($13.8$ seconds) on TextCraft. Measurements use synchronized, unbatched bf16 decoding on one
A100 and exclude prompt construction, parsing, and environment execution.

\subsubsection{Shared and model-exclusive failures}
\label{app:errors}

We compare the two generated plans on each primary test task, using the same
Qwen3-8B executor and evaluation protocol as \Cref{tab:main}. A shared failure
means both plans fail; a diffusion-only failure means the diffusion plan
fails and the AR plan succeeds, and conversely for an AR-only failure.
Of $134$ ALFWorld tasks, $50$ are shared failures, $25$ diffusion-only
failures, and $23$ AR-only failures. The corresponding counts on $100$
TextCraft tasks are $21$, $16$, and $13$ (\Cref{tab:agreement}).

\paragraph{Shared difficulties and differences by task.}
\Cref{tab:failure-families} breaks down ALFWorld outcomes by task family.
Both planners fail on $12/18$ light-inspection tasks and $10/17$ two-object
placement tasks, compared with $3/24$ simple placement tasks. Their similar
aggregate success rates also conceal differences across task families:
cooling has more diffusion-only failures ($9$ versus $2$), while cleaning
has more AR-only failures ($7$ versus $3$). These are descriptive comparisons
on small task subsets, conditional on the evaluated checkpoints and executions.

\begin{table}[H]
\centering
\caption{\textbf{Paired ALFWorld generation failures by task family.}
Each task contributes to at most one failure column; tasks where both
planners succeed are omitted from those columns but included in the total.}
\label{tab:failure-families}
\small
\setlength{\tabcolsep}{6pt}
\begin{tabular}{lrrrr}
\toprule
Task family & Tasks & Both fail & Only diffusion fails & Only AR fails \\
\midrule
Light inspection & $18$ & $12$ & $3$ & $0$ \\
Simple placement & $24$ & $3$ & $2$ & $5$ \\
Clean then place & $31$ & $12$ & $3$ & $7$ \\
Cool then place & $21$ & $6$ & $9$ & $2$ \\
Heat then place & $23$ & $7$ & $4$ & $6$ \\
Two-object placement & $17$ & $10$ & $4$ & $3$ \\
\midrule
Total & $134$ & $50$ & $25$ & $23$ \\
\bottomrule
\end{tabular}
\end{table}

\paragraph{Stopping reasons when only one planner fails.}
\Cref{tab:exclusive-stop-reasons} reports the executor's recorded stopping
reason on tasks where exactly one generated plan succeeds. Failure at a leaf step
is the most frequent reason for both planners in both environments. The
invalid-action limit also occurs frequently in ALFWorld ($9/25$ diffusion-only
and $8/23$ AR-only failures). Completing the plan without terminal success
occurs in $4/25$ diffusion-only ALFWorld failures and none of the $23$ AR-only
failures; in TextCraft it occurs for both planners ($3/16$ and $1/13$).
Thus, these model-exclusive failed tasks mostly share stopping-reason
categories. The labels describe how execution stops, and the stopping-reason
breakdown covers tasks where exactly one planner succeeds.

\begin{table}[H]
\centering
\caption{\textbf{Stopping reasons on tasks failed by only one planner.}
Columns identify the failing planner; the other planner succeeds on each
counted task. Counts cover every model-exclusive failure.}
\label{tab:exclusive-stop-reasons}
\small
\setlength{\tabcolsep}{6pt}
\begin{tabular}{lrrrr}
\toprule
& \multicolumn{2}{c}{ALFWorld} & \multicolumn{2}{c}{TextCraft} \\
\cmidrule(lr){2-3}\cmidrule(lr){4-5}
Recorded stopping reason & Diffusion & AR & Diffusion & AR \\
\midrule
Failure at a leaf step & $10$ & $13$ & $10$ & $7$ \\
No-progress limit & $2$ & $2$ & $2$ & $3$ \\
Plan completed without terminal success & $4$ & $0$ & $3$ & $1$ \\
Invalid-action limit & $9$ & $8$ & $1$ & $2$ \\
\midrule
Total & $25$ & $23$ & $16$ & $13$ \\
\bottomrule
\end{tabular}
\end{table}

\paragraph{Failure locations across both planners.}
We analyze the $148$ ALFWorld and $71$ TextCraft failed executions from the
primary generation evaluation. These diagnostics pool failures from both
planners, so a task failed by both contributes two executions. A failing leaf
step is identified in $106$ ALFWorld and $14$ TextCraft executions; the
remaining $42$ and $57$ failures have no identified failing leaf step in this
diagnostic. This coverage differs from the reference-assisted judge's
repair-region annotations used in \Cref{tab:detector}.

\begin{table}[H]
\centering
\caption{\textbf{Where failed executions stop.} Action and reference-comparison
counts cover failures with an identified leaf step and describe the
observed failure location.}
\label{tab:execution-errors}
\small
\setlength{\tabcolsep}{7pt}
\begin{tabular}{lrr}
\toprule
Diagnostic & ALFWorld & TextCraft \\
\midrule
All failed executions & $148$ & $71$ \\
Identified failing leaf step & $106$ & $14$ \\
Unidentified failing leaf step & $42$ & $57$ \\
\midrule
\multicolumn{3}{l}{Action at the identified failing leaf step} \\
\texttt{take} & $89$ & $0$ \\
\texttt{fetch} & $0$ & $12$ \\
\texttt{goto} & $7$ & $0$ \\
\texttt{put} & $7$ & $0$ \\
\texttt{clean}, \texttt{cool}, \texttt{heat} & $1$ each & $0$ \\
\texttt{craft} & $0$ & $2$ \\
\midrule
\multicolumn{3}{l}{Identified failing leaf step compared with the reference} \\
Identical & $103$ & $9$ \\
Different & $2$ & $5$ \\
Absent from the reference & $1$ & $0$ \\
\bottomrule
\end{tabular}
\end{table}

Object acquisition accounts for $89/106$ identified failures in ALFWorld and
$12/14$ in TextCraft. The failing leaf step matches the reference in
$103/106$ and $9/14$ cases. These counts describe agreement at the failing
leaf step. The action distribution is conditional on identifying that step,
covering $106/148$ ALFWorld and $14/71$ TextCraft failed executions.

\subsubsection{Worked repair and preservation}
\label{app:preservation}

\Cref{fig:teaser} illustrates reference-assisted repair of an ALFWorld plan.
The selected region includes a leaf step and its associated reasoning and
intent fields; executable steps outside the region remain unchanged.

\input{figs/fig1_overview.tex}

We analyze $104$ ALFWorld and $38$ TextCraft failed executions with at least
one completed step. \Cref{fig:failure-progress} reports the pooled fraction of
executor tokens consumed before failure: $56\%$ on ALFWorld and $82\%$ on
TextCraft. After reference-assisted repair, the fractions of diffusion and AR
plans that parse and preserve every completed step unchanged are $40\%$ and
$38\%$ on ALFWorld, and $55\%$ and $74\%$ on TextCraft
(\Cref{fig:repair-preservation}). These are measurements of plan-text
preservation; repaired plans are executed from the initial state.
The fraction of repairs that parse and preserve completed steps is higher for
AR on TextCraft and similar across planners on ALFWorld.

\input{figs/fig_repair_preservation.tex}

\subsection{Sensitivity analyses}
\label{app:sensitivity}

\subsubsection{Representation and executor}
\label{app:nlsurface}
\label{app:sonnet}

\paragraph{Alternative plan representation.}
We retrain both planners in each environment using an alternative structured
plan representation with phase headings and numbered prose steps in place of
structural delimiters. We refer to these two structured formats as prose and
delimited plans, respectively. Task goals, executable fields, task splits, and
model-specific decoding methods remain unchanged. For each representation,
Qwen3-8B and Claude Sonnet 5 execute the same stored plans on the same tasks
with the same budgets.

\begin{table}[H]
\centering
\caption{\textbf{Generation under two representations and executors.} Entries are
task success rates. Within each representation, both executors receive identical
stored plan text.}
\label{tab:nlgen}
\small
\setlength{\tabcolsep}{4.8pt}
\begin{tabular}{llcccc}
\toprule
& & \multicolumn{2}{c}{Qwen3-8B executor} & \multicolumn{2}{c}{Sonnet 5 executor} \\
\cmidrule(lr){3-4}\cmidrule(lr){5-6}
Environment & Planner & Delimited & Prose & Delimited & Prose \\
\midrule
\multirow{2}{*}{ALFWorld}
 & \diffp & $44.0\%$ & $61.9\%$ & $96.3\%$ & $95.5\%$ \\
 & \arp   & $45.5\%$ & $67.9\%$ & $94.8\%$ & $94.8\%$ \\
\midrule
\multirow{2}{*}{TextCraft}
 & \diffp & $63.0\%$ & $69.0\%$ & $94.0\%$ & $94.0\%$ \\
 & \arp   & $66.0\%$ & $78.0\%$ & $94.0\%$ & $96.0\%$ \\
\bottomrule
\end{tabular}
\end{table}

With Qwen3-8B, prose plans improve task success in both environments
(\Cref{tab:nlgen}). The increase on ALFWorld is from $44.0\%$ to $61.9\%$
for \diffp and from $45.5\%$ to $67.9\%$ for \arp. Holding plan text fixed,
Sonnet 5 achieves higher success than Qwen3-8B for both planners and
representations, showing that executor choice also affects task success.

For reference-assisted repair of prose plans, we evaluate only failures for
which the judge selects a usable repair region. This subset differs from the
primary repair population, which retains failures without a selected region.
\Cref{tab:nlrep} reports counts before and after selection. As in the primary
evaluation, a task failed by both planners contributes two distinct failed plans.

\begin{table}[H]
\centering
\caption{\textbf{Reference-assisted repair of prose plans.} ``Failures'' gives counts
from \diffp and \arp, in that order; ``Eligible'' counts plans with a selected
repair region. Both planners receive the same failed plans and regions. Repair
entries report task success after repair and restart.}
\label{tab:nlrep}
\small
\setlength{\tabcolsep}{4.5pt}
\begin{tabular}{llrrrr}
\toprule
Environment & Executor & Failures & Eligible & \diffp repair & \arp repair \\
\midrule
ALFWorld  & Qwen3-8B & $51+43=94$ & $46+43=89$ & $19/89\;(21.3\%)$ & $49/89\;(55.1\%)$ \\
TextCraft & Qwen3-8B & $31+22=53$ & $14+6=20$  & $1/20\;(5.0\%)$  & $6/20\;(30.0\%)$ \\
ALFWorld  & Sonnet 5 & $6+7=13$   & $13$        & $4/13\;(30.8\%)$ & $6/13\;(46.2\%)$ \\
TextCraft & Sonnet 5 & $6+4=10$   & $5$         & $2/5\;(40.0\%)$  & $1/5\;(20.0\%)$ \\
\bottomrule
\end{tabular}
\end{table}

With Qwen3-8B, \arp achieves higher prose-plan repair success in both
environments. With Sonnet 5, the ordering differs across environments, although
the TextCraft comparison includes only five failed plans. Repair outcomes
depend on both representation and executor.

\subsubsection{Repair from execution history}
\label{app:detector}

We prompt the Qwen3-8B executor separately to act as the history-based failure
detector. It receives the goal, current plan, execution
history, and environment feedback, without a reference plan. In one call, it selects
one or more step identifiers and supplies repair feedback, or abstains.
For this comparison, repair feedback is held fixed across selection methods.
\Cref{app:envs} defines the evaluated subset.

\begin{table}[H]
\centering
\caption{\textbf{Repair under reference-assisted and history-based
region selection.} Entries are task success rates after repair. Paired
columns use the same failed plans and repair feedback; abstentions remain in the history-based
denominator.}
\label{tab:detector}
\small
\setlength{\tabcolsep}{6pt}
\begin{tabular}{lcccc}
\toprule
& \multicolumn{2}{c}{\diffp} & \multicolumn{2}{c}{\arp} \\
\cmidrule(lr){2-3}\cmidrule(lr){4-5}
Environment & Reference-assisted & History-based & Reference-assisted & History-based \\
\midrule
ALFWorld ($n=126$)  & $31.7\%$ & $14.3\%$ & $32.5\%$ & $17.5\%$ \\
TextCraft ($n=49$) & $22.4\%$ & $24.5\%$ & $22.4\%$ & $30.6\%$ \\
\midrule
Pooled ($n=175$)   & $29.1\%$ & $17.1\%$ & $29.7\%$ & $21.1\%$ \\
\bottomrule
\end{tabular}
\end{table}

Relative to reference-assisted selection, history-based selection lowers repair
success on ALFWorld and increases it on TextCraft for both planners.

\subsubsection{Natural Plan generation and repair}
\label{app:np}

\input{figs/fig_np.tex}

We evaluate \texttt{Dream-org/DreamReasoner-8B} and
\texttt{Qwen/Qwen3-8B} without task-specific fine-tuning. Generation uses
$1{,}000$ tasks with reference plans that pass the benchmark validators,
comprising $400$ trip-planning, $300$ meeting-scheduling, and $300$
calendar-scheduling tasks. Reasoning is disabled for both checkpoints;
\diffp uses block size $4$ for generation and $256$ for repair.

Each planner receives the benchmark's \texttt{prompt\_5shot} field, containing
the task instruction, five released demonstrations, the test task, and the
final \texttt{SOLUTION:} prefix. For repair, we blank one or two regions of the
reference plan, which follows the benchmark's structured natural-language
format. Both planners receive the same preserved prefix, suffix, repair feedback,
and approximate replacement length, and decode only the missing text.

We construct five repair candidates per reference plan. Three constructions
remove a single region, either a random step, a phrase within a step, or a
constraint-determined value. The other two remove a pair of regions, either
mutually constrained regions or regions in different parts of the plan. We
exclude candidates whose target text occurs verbatim in the visible plan or
prompt.

Of the $5{,}000$ candidates, we retain $2{,}661$ for which restoring the reference
text passes the validator and filling every blank with a fixed nonsensical
string fails. This invalid-replacement control therefore has $0\%$ acceptance
on the retained examples by construction. A separate do-nothing control removes
all blank markers without inserting text. Its validator acceptance falls from
$46.3\%$ on all candidates to $1.3\%$ on the retained examples. All repair
results below use the retained population.

The benchmark validator checks each completed plan directly, without an
executor. Two LLM judges also assess completed plans, receiving only the task
and candidate plan, without the reference plan or validator output
(\Cref{app:npjudge}). For repair, we additionally report exact replacement
recovery, which requires all replacements to match the reference text after
Unicode normalization, whitespace collapse, and removal of leading and trailing
punctuation.

\Cref{tab:np} reports validator and judge acceptance. Exact replacement
recovery is $22.7\%$ for \diffp and $5.0\%$ for \arp.

\begin{table}[H]
\centering
\caption{\textbf{Natural Plan repair by domain and construction.} Entries are
validator acceptance rates on the $2{,}661$ retained examples. Calendar references
contain only a few words specifying a meeting day and time; no calendar
candidates pass the repair filters, so this domain is evaluated only for generation.}
\label{tab:npbreakdown}
\scriptsize
\begin{minipage}[t]{0.42\textwidth}
\centering
\setlength{\tabcolsep}{2.5pt}
\begin{tabular}{lrrr}
\toprule
Domain & $n$ & \diffp & \arp \\
\midrule
Trip planning & $1{,}261$ & $93.3\%$ & $35.7\%$ \\
Meeting scheduling & $1{,}400$ & $18.1\%$ & $19.2\%$ \\
Calendar scheduling & $0$ & -- & -- \\
\bottomrule
\end{tabular}
\end{minipage}\hfill
\begin{minipage}[t]{0.56\textwidth}
\centering
\setlength{\tabcolsep}{3.5pt}
\begin{tabular}{lrrr}
\toprule
Construction & $n$ & \diffp & \arp \\
\midrule
Random step & $545$ & $52.7\%$ & $40.7\%$ \\
Random word span & $491$ & $65.0\%$ & $28.5\%$ \\
Constraint assignment & $482$ & $69.5\%$ & $47.9\%$ \\
Causally linked regions & $541$ & $34.2\%$ & $0.7\%$ \\
Compound nonlocal regions & $602$ & $50.5\%$ & $20.3\%$ \\
\bottomrule
\end{tabular}
\end{minipage}
\end{table}

The validator and judges agree on the model ordering within each operation,
despite their different acceptance criteria and absolute success rates.
The aggregate repair advantage is concentrated in trip planning, while
\diffp has higher validator acceptance under each of the five constructions. A percentile bootstrap with $50{,}000$ resamples clustered by the
$696$ retained source tasks gives a $95\%$ interval of $[23.8,29.6]$ percentage
points for the aggregate \diffp-minus-\arp difference. Each resample draws source
tasks with replacement and includes all their repair examples.
\input{figs/fig_np_example.tex}

\begin{samepage}
\subsubsection{Long-horizon generation in ScienceWorld}
\label{app:sw}

The ScienceWorld checkpoints are trained separately on simulator-provided
reference plans \citep{scienceworld}. These references average $27.1$ leaf
steps, compared with $5.8$ for ALFWorld and $9.8$ for TextCraft. We evaluate
$825$ held-out generations per planner and group them by reference-plan length.\par
\end{samepage}

\input{figs/fig_sw.tex}

Of $758$ rejected generations across both planners, $754$ reach the generation limit without closing the plan
hierarchy; the remaining four omit a subgoal body. Rejected outputs also contain
repeated actions, with a median duplicated action--argument fraction of $93.3\%$
for \diffp and $40.0\%$ for \arp.
Among parser-rejected plans, the dominant failure is incomplete serialization, accompanied by
extensive repetition, especially for diffusion.

For a nonempty plan with $\leafcount$ leaf steps, let
$(\plantool_\ell,\toolargs_\ell)$ denote the action--argument pair at step
$\ell\in[\leafcount]$ after flattening the hierarchy. The duplicated fraction is
\[
\frac{\left|\left\{\ell\in[\leafcount]:\ \exists k\in[\leafcount]\setminus\{\ell\},\ (\plantool_k,\toolargs_k)=(\plantool_\ell,\toolargs_\ell)\right\}\right|}{\leafcount}.
\]
This counts every occurrence of a repeated pair, including its first occurrence.

\subsection{Implementation details and prompts}
\label{app:hparams}

\subsubsection{Training and decoding}
\label{app:decode}

On ALFWorld and TextCraft, we fully fine-tune the checkpoints specified in
\Cref{sec:setup} with AdamW and bf16 weights. Each environment-specific
checkpoint is trained jointly on generation and repair, with both objectives
present in a single run. An epoch visits each generation and repair example
once. We use one training seed ($42$); reported evaluation
uncertainty does not include variation across training seeds. Activation
checkpointing and Fully Sharded Data Parallel (FSDP) with full sharding and CPU parameter offloading support full
fine-tuning. The planners share task
splits, training examples, and evaluation interfaces (\Cref{tab:hparams}).
Native reasoning mode is disabled for both planners. The plan's
\texttt{[THINK]} field remains part of the supervised structured output.

For diffusion decoding, $\confthr$ is the base confidence threshold. Repair
uses DreamReasoner's RelaxedConfidence procedure \citep{dreamreasoner}, with
threshold floor $\confmin$, neighbour-support cutoff $\supportthr$, and
neighbourhood radius $\supportradius$.

\begin{table}[t]
\centering
\caption{\textbf{ALFWorld and TextCraft training and decoding settings.} Block weights follow the
listed block-size order.}
\label{tab:hparams}
\small
\setlength{\tabcolsep}{5pt}
\begin{tabular}{@{}>{\raggedright\arraybackslash}p{0.25\linewidth}
                  >{\raggedright\arraybackslash}p{0.35\linewidth}
                  >{\raggedright\arraybackslash}p{\dimexpr0.40\linewidth-20pt\relax}@{}}
\toprule
Setting & Plan generation & Plan repair \\
\midrule
Training endpoint & \multicolumn{2}{l}{Convergence of development-set performance} \\
Learning rate & $10^{-5}$ & $10^{-5}$ \\
Batch $\times$ gradient accumulation & $1\times16$ & $1\times16$ \\
Learning-rate warmup & \multicolumn{2}{l}{$5\%$ of optimization steps} \\
Maximum sequence length & \multicolumn{2}{l}{$4{,}096$ tokens at training and inference} \\
\addlinespace
\diffp block sizes & $4,8,16,32,64,$\newline $128,256,512,1024$ & $32,64,128,256,2048,4096$ \\
\diffp block weights & $(24,20,16,12,8,6,5,4,3)$ & $(2,3,4,6,10,24)$ \\
\diffp inference & Block-wise denoising, $\blocksize=4$, $\confthr=0.95$ & Full-canvas infilling, $\confthr=0.95$, $\confmin=\supportthr=0.7$, $\supportradius=4$ \\
\arp inference & Left-to-right with key--value cache & Left-to-right replacement generation using FIM serialization \\
\addlinespace
Executor & \multicolumn{2}{l}{Qwen3-8B, temperature $0.6$, top-$p$ $0.95$, one environment action per call} \\
\bottomrule
\end{tabular}
\end{table}

Diffusion repair allocates $2|\repairregion|$ positions in total across the selected
spans. Each span has its own padding, and its replacement ends before the first
\texttt{[delete]} token in its canvas segment. Across environment-specific
settings, a diffusion repair forward pass commits $7.7$--$18.3$ tokens; the AR decoder
generates one token per forward pass using a key--value cache.

\subsubsection{Executor behavior}

The primary executor is Qwen3-8B without task-specific fine-tuning. Each execution call
receives the goal, current observation, complete plan, execution
history, and action grammar, and generates one environment action. The executor
strictly follows the planned actions and arguments in leaf-step order. Execution stops
on success, environment failure, an invalid-action or no-progress limit,
$60$ planned leaf steps, or $1{,}000$ episode tokens.
Parser rejection and execution ending without task success count as failure.
The executor comparison uses Claude Sonnet 5
(\texttt{us.anthropic.claude-sonnet-5}) with the same inputs and budgets and
default model settings.

\subsubsection{Plan representation and prompts}
\label{app:format}
\label{app:prompts}

The following templates use the structured plan representation in
\Cref{def:plan}. Environment-specific system instructions define the available
actions and required fields, delimiters, and nesting. The trained planners
receive no demonstrations. Repair feedback and region selection follow
\Cref{sec:repair}.

\noindent\begin{minipage}[t]{0.48\textwidth}
\textbf{Generation.}\par\smallskip
{\fontsize{7.5}{9}\selectfont\ttfamily\raggedright
System: Return one complete hierarchical plan for <environment> in the supplied schema. Each leaf contains an action, arguments, and expected observation.\par
User:\par
Goal: <task goal>\par
Initial observation: <observation>\par
Available actions: <action grammar>\par
Return only the structured plan.\par}
\end{minipage}\hfill
\begin{minipage}[t]{0.48\textwidth}
\textbf{Repair.}\par\smallskip
{\fontsize{7.5}{9}\selectfont\ttfamily\raggedright
The plan below failed for the same goal.\par
Selected region: <identifier>; Reason: <failure-detector explanation>\par
Expected: <plan prediction>; Observed: <environment observation>\par
Plan: <prefix> <editable region> <suffix>\par
Return only corrected content for the editable region; preserve all other text.\par}
\end{minipage}

\medskip
\paragraph{History-based failure detector.}
In this prompt, ``execution trace'' denotes execution history, and
``Failure feedback'' denotes environment feedback.

{\fontsize{7.5}{9}\selectfont\ttfamily\raggedright
System: Given a goal, structured plan, and execution trace, select the smallest
set of exact step identifiers inconsistent with the observed feedback. Do not
repair; abstain if the trace identifies no defective step.\par
User: Goal: <task goal>; Current plan: <structured plan with step identifiers>\par
Execution history: <ordered actions, expected observations, and observed outcomes>;
Failure feedback: <latest environment observation>\par
Return exactly: REPLAN\_NEEDED: <yes or no>; STEP\_IDS: <comma-separated step
identifiers; empty if no>; REASON: <brief explanation grounded only in the
execution trace>.\par}

\noindent\begin{minipage}{\linewidth}
\raggedright
\paragraph{Natural Plan judges.}
\label{app:npjudge}
We query GPT-5 (\texttt{gpt-5}) with medium reasoning effort and Claude Sonnet 5
(\texttt{us.anthropic.claude-sonnet-5}) with default model settings, once per
candidate, using the following prompt.

{\fontsize{7.5}{9}\selectfont\ttfamily\raggedright
System: Check the candidate plan against every task requirement step by step.
Return accurate only if it is feasible and satisfies all constraints. Do not
use or infer a reference answer.\par
User: TASK: <task requirements>; CANDIDATE PLAN: <completed plan>\par
Return exactly: REASON: <brief explanation>; ACCURATE: <0 or 1>.\par}
\end{minipage}\par

%% file: figs/tab_training_data.tex
\begin{table}[H]
\centering
\caption{\textbf{Task splits and repair-training examples.} Task counts precede reference-plan execution filtering. Repair examples combine
failed planner executions and structured perturbations of verified plans.}
\label{tab:datasets}
\small
\setlength{\tabcolsep}{5.5pt}
\begin{tabular}{lrrrrrrr}
\toprule
& \multicolumn{4}{c}{Task split} & \multicolumn{3}{c}{Repair-training examples} \\
\cmidrule(lr){2-5}\cmidrule(lr){6-8}
Environment & train-gen & train-rep & dev & test & real & synthetic & total \\
\midrule
ALFWorld  & $2{,}665$ & $888$   & $140$ & $134$ & $699$ & $763$   & $1{,}462$ \\
TextCraft & $3{,}824$ & $1{,}268$ & $50$  & $100$ & $267$ & $1{,}661$ & $1{,}928$ \\
\bottomrule
\end{tabular}
\end{table}

%% file: figs/fig3_results.tex
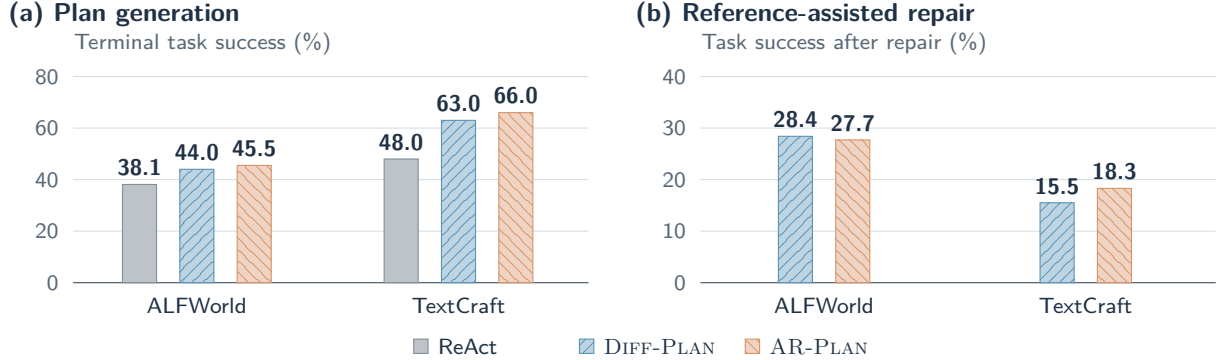
\begin{figure}[t]
\centering
\resizebox{\linewidth}{!}{\input{figure_assets/3_results.tikz}}
\caption{\textbf{Observed success in the two interactive benchmarks.} (a) Task success after plan generation, on $134$ ALFWorld and $100$ TextCraft tasks. (b) Task success after reference-assisted repair, on $148$ and $71$ pooled failed plans. Diffusion--AR differences are at most three percentage points; higher is better.}
\label{fig:main}
\end{figure}

%% file: figure_assets/3_results.tikz
\begin{tikzpicture}[x=1cm,y=1cm,fp text]
\path (0,-0.95) rectangle (13.8,3.13);
\begin{scope}[xshift=0cm]
\node[fp title] at (0.02,2.99) {(a) Plan generation};
\node[fp label,anchor=west] at (0.72,2.67) {Terminal task success (\%)};
\draw[fp grid] (0.75,0.0000)--(6.50,0.0000);
\node[fp label,anchor=east] at (0.63,0.0000) {0};
\draw[fp grid] (0.75,0.5750)--(6.50,0.5750);
\node[fp label,anchor=east] at (0.63,0.5750) {20};
\draw[fp grid] (0.75,1.1500)--(6.50,1.1500);
\node[fp label,anchor=east] at (0.63,1.1500) {40};
\draw[fp grid] (0.75,1.7250)--(6.50,1.7250);
\node[fp label,anchor=east] at (0.63,1.7250) {60};
\draw[fp grid] (0.75,2.3000)--(6.50,2.3000);
\node[fp label,anchor=east] at (0.63,2.3000) {80};
\draw[fp axis] (0.75,0)--(6.50,0);
\node[fp label,anchor=north,text=ink] at (2.13,-0.13) {ALFWorld};
\path[fp baseline bar] (1.3000,0) rectangle (1.6800,1.0954);
\node[fp value,anchor=south] at (1.4900,1.1554) {38.1};
\path[fp diff bar] (1.9400,0) rectangle (2.3200,1.2650);
\node[fp value,anchor=south] at (2.1300,1.3250) {44.0};
\path[fp ar bar] (2.5800,0) rectangle (2.9600,1.3081);
\node[fp value,anchor=south] at (2.7700,1.3681) {45.5};
\node[fp label,anchor=north,text=ink] at (5.05,-0.13) {TextCraft};
\path[fp baseline bar] (4.2200,0) rectangle (4.6000,1.3800);
\node[fp value,anchor=south] at (4.4100,1.4400) {48.0};
\path[fp diff bar] (4.8600,0) rectangle (5.2400,1.8112);
\node[fp value,anchor=south] at (5.0500,1.8712) {63.0};
\path[fp ar bar] (5.5000,0) rectangle (5.8800,1.8975);
\node[fp value,anchor=south] at (5.6900,1.9575) {66.0};
\end{scope}
\begin{scope}[xshift=7.0cm]
\node[fp title] at (0.02,2.99) {(b) Reference-assisted repair};
\node[fp label,anchor=west] at (0.72,2.67) {Task success after repair (\%)};
\draw[fp grid] (0.75,0.0000)--(6.50,0.0000);
\node[fp label,anchor=east] at (0.63,0.0000) {0};
\draw[fp grid] (0.75,0.5750)--(6.50,0.5750);
\node[fp label,anchor=east] at (0.63,0.5750) {10};
\draw[fp grid] (0.75,1.1500)--(6.50,1.1500);
\node[fp label,anchor=east] at (0.63,1.1500) {20};
\draw[fp grid] (0.75,1.7250)--(6.50,1.7250);
\node[fp label,anchor=east] at (0.63,1.7250) {30};
\draw[fp grid] (0.75,2.3000)--(6.50,2.3000);
\node[fp label,anchor=east] at (0.63,2.3000) {40};
\draw[fp axis] (0.75,0)--(6.50,0);
\node[fp label,anchor=north,text=ink] at (2.13,-0.13) {ALFWorld};
\path[fp diff bar] (1.6200,0) rectangle (2.0000,1.6330);
\node[fp value,anchor=south] at (1.8100,1.6930) {28.4};
\path[fp ar bar] (2.2600,0) rectangle (2.6400,1.5927);
\node[fp value,anchor=south] at (2.4500,1.6527) {27.7};
\node[fp label,anchor=north,text=ink] at (5.05,-0.13) {TextCraft};
\path[fp diff bar] (4.5400,0) rectangle (4.9200,0.8912);
\node[fp value,anchor=south] at (4.7300,0.9512) {15.5};
\path[fp ar bar] (5.1800,0) rectangle (5.5600,1.0522);
\node[fp value,anchor=south] at (5.3700,1.1122) {18.3};
\end{scope}
\path[fp baseline bar] (4.55,-0.79) rectangle (4.71,-0.63);
\node[fp label,anchor=west,text=ink] at (4.78,-0.71) {ReAct};
\path[fp diff bar] (6.40,-0.79) rectangle (6.56,-0.63);
\node[fp label,anchor=west,text=ink] at (6.63,-0.71) {\diffp};
\path[fp ar bar] (8.25,-0.79) rectangle (8.41,-0.63);
\node[fp label,anchor=west,text=ink] at (8.48,-0.71) {\arp};
\end{tikzpicture}

%% file: figs/fig1_overview.tex
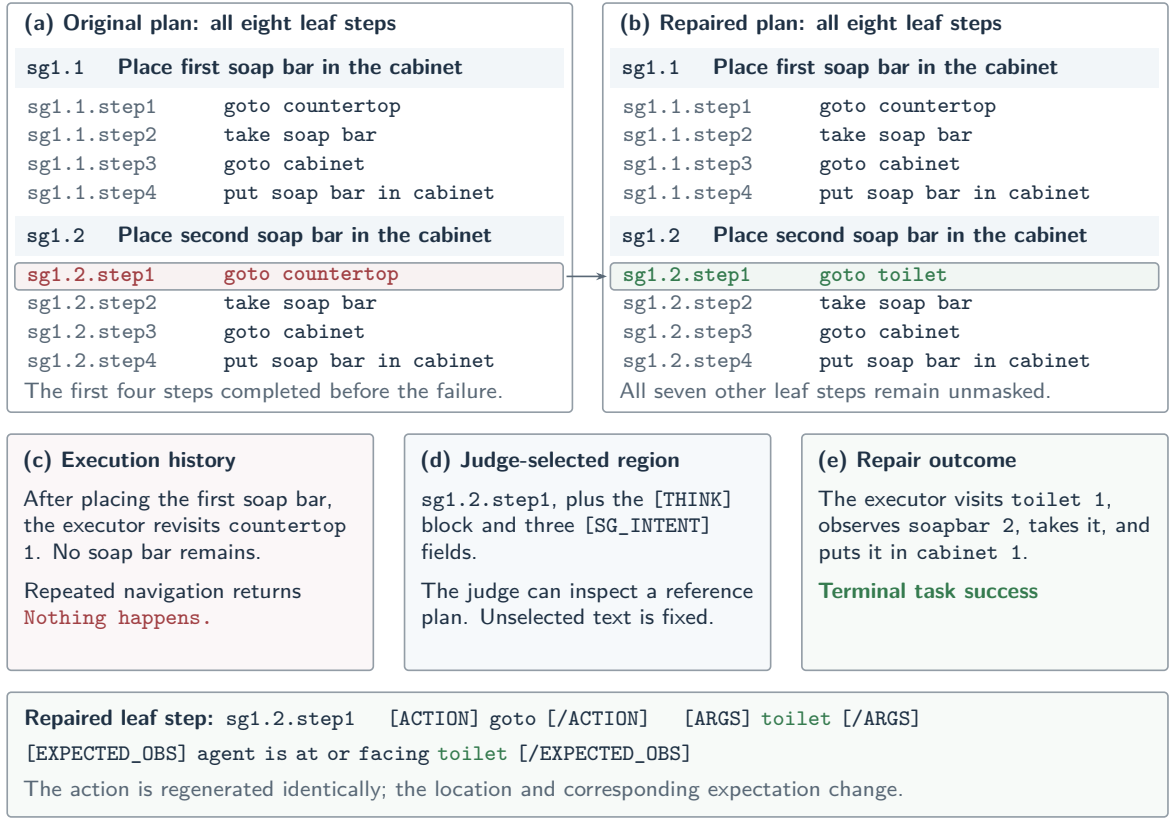
\begin{figure}[t]
\centering
\resizebox{0.94\linewidth}{!}{\input{figure_assets/1_overview.tikz}}
\caption{\textbf{Plan repair in ALFWorld.}
In this training example, a reference-assisted judge selects one leaf step,
the reasoning block, and three intent fields for regeneration. Infilling changes
\texttt{countertop} to \texttt{toilet} and updates the expected observation.
The other seven leaf steps remain fixed. The repaired plan succeeds.}
\label{fig:teaser}
\end{figure}

%% file: figure_assets/1_overview.tikz
\begin{tikzpicture}[x=1cm,y=1cm,text=ink,
 font=\sffamily\fontsize{7.5}{9.0}\selectfont,
 pp border/.style={draw=muted!65,solid,line width=.55pt,rounded corners=1.4pt},
 pp card/.style={pp border,fill=white,anchor=north west,inner sep=0pt},
 pp text/.style={anchor=west,inner sep=0pt},
 pp title/.style={pp text,font=\sffamily\bfseries\fontsize{7.5}{9.0}\selectfont},
 pp code/.style={pp text,font=\ttfamily\fontsize{7.5}{9.0}\selectfont},
 pp note/.style={pp text,text=muted},
 pp arrow/.style={-{Stealth[length=1.3mm,width=.9mm]},draw=muted,line width=.55pt}]
\definecolor{pprepair}{RGB}{53,123,79}
\node[pp title] at (0,0) {ALFWorld: find two soapbar and put them in cabinet};

\begin{scope}[yshift=.27cm]
\foreach \x/\name/\title in {0/before/{(a) Original plan: all eight leaf steps},7.06/after/{(b) Repaired plan: all eight leaf steps}} {
 \node[pp card,minimum width=6.70cm,minimum height=4.85cm] (\name) at (\x,-.63) {};
 \node[pp title] at (\x+.20,-.90) {\title};
 \path[fill=diffcol!6] (\x+.10,-1.18) rectangle (\x+6.60,-1.65);
 \node[pp title] at (\x+.23,-1.415)
 {\texttt{sg1.1} \quad Place first soap bar in the cabinet};
 \foreach \id/\act/\yy in {1/{goto countertop}/1.89,2/{take soap bar}/2.23,3/{goto cabinet}/2.57,4/{put soap bar in cabinet}/2.91} {
  \node[pp code,text=muted] at (\x+.24,-\yy) {sg1.1.step\id};
  \node[pp code] at (\x+2.57,-\yy) {\act};
 }
 \path[fill=diffcol!6] (\x+.10,-3.17) rectangle (\x+6.60,-3.64);
 \node[pp title] at (\x+.23,-3.405)
 {\texttt{sg1.2} \quad Place second soap bar in the cabinet};
 \foreach \id/\act/\yy in {2/{take soap bar}/4.22,3/{goto cabinet}/4.56,4/{put soap bar in cabinet}/4.90} {
  \node[pp code,text=muted] at (\x+.24,-\yy) {sg1.2.step\id};
  \node[pp code] at (\x+2.57,-\yy) {\act};
 }
}
\path[pp border,fill=errorcol!7] (.10,-3.72) rectangle (6.60,-4.04);
\node[pp code,text=errorcol] at (.24,-3.88) {\textbf{sg1.2.step1}};
\node[pp code,text=errorcol] at (2.57,-3.88) {\textbf{goto countertop}};
\path[pp border,fill=pprepair!7] (7.16,-3.72) rectangle (13.66,-4.04);
\node[pp code,text=pprepair] at (7.30,-3.88) {\textbf{sg1.2.step1}};
\node[pp code,text=pprepair] at (9.63,-3.88) {\textbf{goto toilet}};
\draw[pp arrow] (6.64,-3.88) -- (7.12,-3.88);
\node[pp note] at (.20,-5.25) {The first four steps completed before the failure.};
\node[pp note] at (7.26,-5.25) {All seven other leaf steps remain unmasked.};

\foreach \x/\shade in {0/errorcol,4.71/diffcol,9.42/pprepair} {
 \node[pp card,fill=\shade!4,minimum width=4.34cm,minimum height=2.80cm]
 at (\x,-5.74) {};
}
\node[anchor=north west,inner sep=0pt] at (.20,-5.94) {
\begin{minipage}[t]{3.94cm}\raggedright
 \textbf{(c) Execution history}\\[4pt]
 After placing the first soap bar, the executor revisits \texttt{countertop 1}. No soap bar remains.\\[4pt]
 Repeated navigation returns \textcolor{errorcol}{\texttt{Nothing happens.}}
\end{minipage}};
\node[anchor=north west,inner sep=0pt] at (4.91,-5.94) {
\begin{minipage}[t]{3.94cm}\raggedright
 \textbf{(d) Judge-selected region}\\[4pt]
 \texttt{sg1.2.step1}, plus the \texttt{[THINK]} block and three \texttt{[SG\_INTENT]} fields.\\[4pt]
 The judge can inspect a reference plan. Unselected text is fixed.
\end{minipage}};
\node[anchor=north west,inner sep=0pt] at (9.62,-5.94) {
\begin{minipage}[t]{3.94cm}\raggedright
 \textbf{(e) Repair outcome}\\[4pt]
 The executor visits \texttt{toilet 1}, observes \texttt{soapbar 2}, takes it, and puts it in \texttt{cabinet 1}.\\[4pt]
 \textcolor{pprepair}{\textbf{Terminal task success}}
\end{minipage}};

\node[pp card,fill=pprepair!4,text width=13.36cm,inner sep=.20cm] at (0,-8.80) {
 \textbf{Repaired leaf step:} \texttt{sg1.2.step1} \quad
 \texttt{[ACTION] goto [/ACTION]} \quad
 \texttt{[ARGS] \textcolor{pprepair}{toilet} [/ARGS]}\\[3pt]
 \texttt{[EXPECTED\_OBS] agent is at or facing \textcolor{pprepair}{toilet} [/EXPECTED\_OBS]}\\[3pt]
 \textcolor{muted}{The action is regenerated identically; the location and corresponding expectation change.}};
\end{scope}
\end{tikzpicture}

%% file: figs/fig_repair_preservation.tex
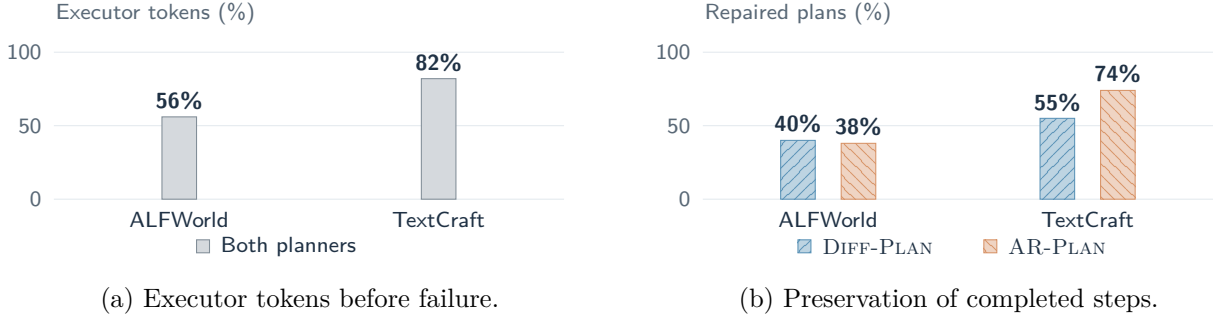
\begin{figure}[htbp]
\centering
\begin{subfigure}[t]{0.48\linewidth}
\centering
\resizebox{\linewidth}{!}{\input{figure_assets/interaction_before_failure.tikz}}
\caption{Executor tokens before failure.}
\label{fig:failure-progress}
\end{subfigure}
\hfill
\begin{subfigure}[t]{0.48\linewidth}
\centering
\resizebox{\linewidth}{!}{\input{figure_assets/repair_preservation.tikz}}
\caption{Preservation of completed steps.}
\label{fig:repair-preservation}
\end{subfigure}
\caption{\textbf{Interaction before failure and preservation after repair.}
We analyze $104$ ALFWorld and $38$ TextCraft failed executions with at least one
completed step. Plans come from both planners and use the Qwen3-8B executor.
(a) Pooled fraction of executor tokens consumed before the recorded failure point.
(b) Fraction of plans after reference-assisted repair that parse and preserve
every completed step, shown separately for each planner. Preservation compares completed-step
text before and after repair; repaired plans are executed from the initial state.}
\label{fig:failure-analysis}
\end{figure}

%% file: figure_assets/interaction_before_failure.tikz
\begin{tikzpicture}[x=1cm,y=1cm,fp text,
 fp label/.style={font=\sffamily\fontsize{7.2}{8.4}\selectfont,text=muted,inner sep=1pt},
 fp value/.style={font=\sffamily\bfseries\fontsize{7.2}{8.4}\selectfont,text=ink,inner sep=1pt}]
\path[use as bounding box] (0,-0.80) rectangle (6.8,2.35);
\node[fp label,anchor=west] at (0.58,2.15) {Executor tokens (\%)};
\foreach \y/\lab in {0/0,0.85/50,1.70/100} {
  \draw[draw=rulecol!65,line width=0.2pt] (0.60,\y)--(6.50,\y);
  \node[fp label,anchor=east] at (0.50,\y) {\lab};
}
\path[fp pooled bar] (1.85,0) rectangle (2.25,0.952);
\path[fp pooled bar] (4.85,0) rectangle (5.25,1.394);
\node[fp value,anchor=south] at (2.05,1.00) {56\%};
\node[fp value,anchor=south] at (5.05,1.44) {82\%};
\node[fp label,text=ink,anchor=north] at (2.05,-0.10) {ALFWorld};
\node[fp label,text=ink,anchor=north] at (5.05,-0.10) {TextCraft};
\path[fp pooled bar] (2.25,-.62) rectangle (2.40,-.47);
\node[fp label,anchor=west,text=ink] at (2.49,-.545) {Both planners};
\end{tikzpicture}

%% file: figure_assets/repair_preservation.tikz
\begin{tikzpicture}[x=1cm,y=1cm,fp text,
 fp label/.style={font=\sffamily\fontsize{7.2}{8.4}\selectfont,text=muted,inner sep=1pt},
 fp value/.style={font=\sffamily\bfseries\fontsize{7.2}{8.4}\selectfont,text=ink,inner sep=1pt}]
\path[use as bounding box] (0,-0.80) rectangle (6.8,2.35);
\node[fp label,anchor=west] at (0.58,2.15) {Repaired plans (\%)};
\foreach \y/\lab in {0/0,0.85/50,1.70/100} {
  \draw[draw=rulecol!65,line width=0.2pt] (0.60,\y)--(6.50,\y);
  \node[fp label,anchor=east] at (0.50,\y) {\lab};
}
\path[fp diff bar] (1.50,0) rectangle (1.90,0.680);
\path[fp ar bar] (2.20,0) rectangle (2.60,0.646);
\path[fp diff bar] (4.50,0) rectangle (4.90,0.935);
\path[fp ar bar] (5.20,0) rectangle (5.60,1.258);
\node[fp value,anchor=south] at (1.70,0.73) {40\%};
\node[fp value,anchor=south] at (2.40,0.70) {38\%};
\node[fp value,anchor=south] at (4.70,0.98) {55\%};
\node[fp value,anchor=south] at (5.40,1.31) {74\%};
\node[fp label,text=ink,anchor=north] at (2.05,-0.10) {ALFWorld};
\node[fp label,text=ink,anchor=north] at (5.05,-0.10) {TextCraft};
\path[fp diff bar] (1.70,-.62) rectangle (1.85,-.47);
\node[fp label,anchor=west,text=ink] at (1.94,-.545) {\diffp};
\path[fp ar bar] (3.85,-.62) rectangle (4.00,-.47);
\node[fp label,anchor=west,text=ink] at (4.09,-.545) {\arp};
\end{tikzpicture}

%% file: figs/fig_np.tex
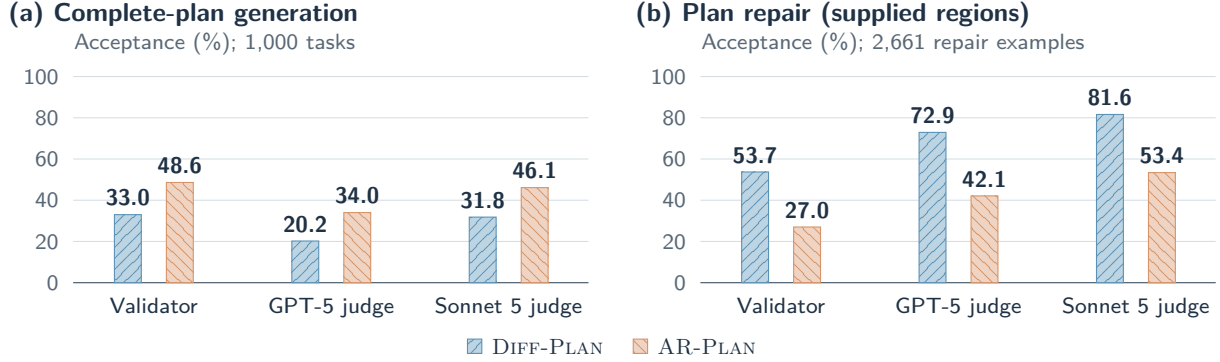
\begin{figure}[htbp]
\centering
\resizebox{\linewidth}{!}{\input{figure_assets/np.tikz}}
\caption{\textbf{Plan generation and repair on Natural Plan.}
Both models are evaluated without task-specific fine-tuning.
Generation uses $1{,}000$ tasks; repair infills supplied regions in $2{,}661$
examples from $696$ tasks. Bars show acceptance rates under the benchmark
validator and two language-model judges. \Cref{app:np} gives the filtering
protocol and complete results.}
\label{fig:np}
\end{figure}

%% file: figure_assets/np.tikz
\begin{tikzpicture}[x=1cm,y=1cm,fp text]
\path (0,-0.95) rectangle (13.8,3.13);
\begin{scope}[xshift=0cm]
\node[fp title] at (0.02,2.99) {(a) Complete-plan generation};
\node[fp label,anchor=west] at (0.72,2.67) {Acceptance (\%); 1,000 tasks};
\draw[fp grid] (0.75,0.0000)--(6.50,0.0000);
\node[fp label,anchor=east] at (0.63,0.0000) {0};
\draw[fp grid] (0.75,0.4600)--(6.50,0.4600);
\node[fp label,anchor=east] at (0.63,0.4600) {20};
\draw[fp grid] (0.75,0.9200)--(6.50,0.9200);
\node[fp label,anchor=east] at (0.63,0.9200) {40};
\draw[fp grid] (0.75,1.3800)--(6.50,1.3800);
\node[fp label,anchor=east] at (0.63,1.3800) {60};
\draw[fp grid] (0.75,1.8400)--(6.50,1.8400);
\node[fp label,anchor=east] at (0.63,1.8400) {80};
\draw[fp grid] (0.75,2.3000)--(6.50,2.3000);
\node[fp label,anchor=east] at (0.63,2.3000) {100};
\draw[fp axis] (0.75,0)--(6.50,0);
\node[fp label,anchor=north,text=ink] at (1.65,-0.13) {Validator};
\path[fp diff bar] (1.2100,0) rectangle (1.5100,0.7590);
\node[fp value,anchor=south] at (1.3600,0.8190) {33.0};
\path[fp ar bar] (1.7900,0) rectangle (2.0900,1.1178);
\node[fp value,anchor=south] at (1.9400,1.1778) {48.6};
\node[fp label,anchor=north,text=ink] at (3.63,-0.13) {GPT-5 judge};
\path[fp diff bar] (3.1900,0) rectangle (3.4900,0.4646);
\node[fp value,anchor=south] at (3.3400,0.5246) {20.2};
\path[fp ar bar] (3.7700,0) rectangle (4.0700,0.7820);
\node[fp value,anchor=south] at (3.9200,0.8420) {34.0};
\node[fp label,anchor=north,text=ink] at (5.61,-0.13) {Sonnet 5 judge};
\path[fp diff bar] (5.1700,0) rectangle (5.4700,0.7314);
\node[fp value,anchor=south] at (5.3200,0.7914) {31.8};
\path[fp ar bar] (5.7500,0) rectangle (6.0500,1.0603);
\node[fp value,anchor=south] at (5.9000,1.1203) {46.1};
\end{scope}
\begin{scope}[xshift=7.0cm]
\node[fp title] at (0.02,2.99) {(b) Plan repair (supplied regions)};
\node[fp label,anchor=west] at (0.72,2.67) {Acceptance (\%); 2,661 repair examples};
\draw[fp grid] (0.75,0.0000)--(6.50,0.0000);
\node[fp label,anchor=east] at (0.63,0.0000) {0};
\draw[fp grid] (0.75,0.4600)--(6.50,0.4600);
\node[fp label,anchor=east] at (0.63,0.4600) {20};
\draw[fp grid] (0.75,0.9200)--(6.50,0.9200);
\node[fp label,anchor=east] at (0.63,0.9200) {40};
\draw[fp grid] (0.75,1.3800)--(6.50,1.3800);
\node[fp label,anchor=east] at (0.63,1.3800) {60};
\draw[fp grid] (0.75,1.8400)--(6.50,1.8400);
\node[fp label,anchor=east] at (0.63,1.8400) {80};
\draw[fp grid] (0.75,2.3000)--(6.50,2.3000);
\node[fp label,anchor=east] at (0.63,2.3000) {100};
\draw[fp axis] (0.75,0)--(6.50,0);
\node[fp label,anchor=north,text=ink] at (1.65,-0.13) {Validator};
\path[fp diff bar] (1.2100,0) rectangle (1.5100,1.2351);
\node[fp value,anchor=south] at (1.3600,1.2951) {53.7};
\path[fp ar bar] (1.7900,0) rectangle (2.0900,0.6210);
\node[fp value,anchor=south] at (1.9400,0.6810) {27.0};
\node[fp label,anchor=north,text=ink] at (3.63,-0.13) {GPT-5 judge};
\path[fp diff bar] (3.1900,0) rectangle (3.4900,1.6767);
\node[fp value,anchor=south] at (3.3400,1.7367) {72.9};
\path[fp ar bar] (3.7700,0) rectangle (4.0700,0.9683);
\node[fp value,anchor=south] at (3.9200,1.0283) {42.1};
\node[fp label,anchor=north,text=ink] at (5.61,-0.13) {Sonnet 5 judge};
\path[fp diff bar] (5.1700,0) rectangle (5.4700,1.8768);
\node[fp value,anchor=south] at (5.3200,1.9368) {81.6};
\path[fp ar bar] (5.7500,0) rectangle (6.0500,1.2282);
\node[fp value,anchor=south] at (5.9000,1.2882) {53.4};
\end{scope}
\path[fp diff bar] (5.15,-0.79) rectangle (5.31,-0.63);
\node[fp label,anchor=west,text=ink] at (5.38,-0.71) {\diffp};
\path[fp ar bar] (7.00,-0.79) rectangle (7.16,-0.63);
\node[fp label,anchor=west,text=ink] at (7.23,-0.71) {\arp};
\end{tikzpicture}

%% file: figs/fig_np_example.tex
\begin{figure}[t]
\centering
\resizebox{\linewidth}{!}{\input{figure_assets/np_example.tikz}}
\caption{\textbf{An example of plan repair on Natural Plan.} Both planners receive the same task, repair feedback, preserved prefix and suffix, and approximate replacement length. The blank is the prefix of a line, so its replacement must join the preserved suffix. The diffusion output restores the reference phrase; the AR output repeats the following line. This example illustrates the within-task comparison in \Cref{fig:np}.}
\label{fig:np-example}
\end{figure}

%% file: figure_assets/np_example.tikz
\begin{tikzpicture}[x=1cm,y=1cm,fp text]
\path (0,-0.1) rectangle (13.8,6.8);
\node[fp title] at (0.02,6.59) {Natural Plan: reconstruct one constraint-determined phrase};
\node[fp card,anchor=north west,text width=13.42cm] (task) at (0,6.25) {
\textbf{Task.} Visit 4 European cities in 17 days on direct flights: Stockholm 2 days (conference days 1--2),
Dubrovnik 7 days, Vienna 5 days, Mykonos 6 days (friends there days 12--17).
Direct flights: Dubrovnik--Vienna, Stockholm--Vienna, Vienna--Mykonos, Stockholm--Dubrovnik.\\[3pt]
\textbf{Repair feedback.} A short phrase inside one step states a value that the task's constraints do not permit.};
\node[fp card,anchor=north west,fill=white,text width=13.42cm] (context) at (0,4.58) {
\textbf{Plan supplied to both planners.} One region is blanked; the remainder is stated correct.\\[4pt]
{\ttfamily\fontsize{7.4}{10}\selectfont
Day 1-2: Arriving in Stockholm and visit Stockholm for 2 days.\\
Day 2: Fly from Stockholm to Dubrovnik.\quad Day 2-8: Visit Dubrovnik for 7 days.\\
Day 8: Fly from Dubrovnik to Vienna.\quad Day 8-12: Visit Vienna for 5 days.\\
\colorbox{black!8}{\strut\hspace{3.9cm}} Mykonos.\quad {\sffamily(about 30 characters, 6 words)}\\
Day 12-17: Visit Mykonos for 6 days.}};
\node[fp card,anchor=north west,text width=6.24cm,draw=diffcol!40,fill=diffcol!4] (diff) at (0,1.94) {
\textbf{\diffp: exact replacement}\\[4pt]
\texttt{Day 12: Fly from Vienna to}\\[4pt]
The inserted phrase joins the preserved word\\\texttt{Mykonos.} to form the required flight.};
\node[fp card,anchor=north west,text width=6.24cm,draw=arcol!40,fill=arcol!4] (ar) at (7.18,1.94) {
\textbf{\arp: incorrect replacement}\\[4pt]
\texttt{Day 12-17: Visit Mykonos for 6 days.}\\[4pt]
The inserted phrase repeats the following line\\and leaves an extra \texttt{Mykonos.}};
\end{tikzpicture}

%% file: figs/fig_sw.tex
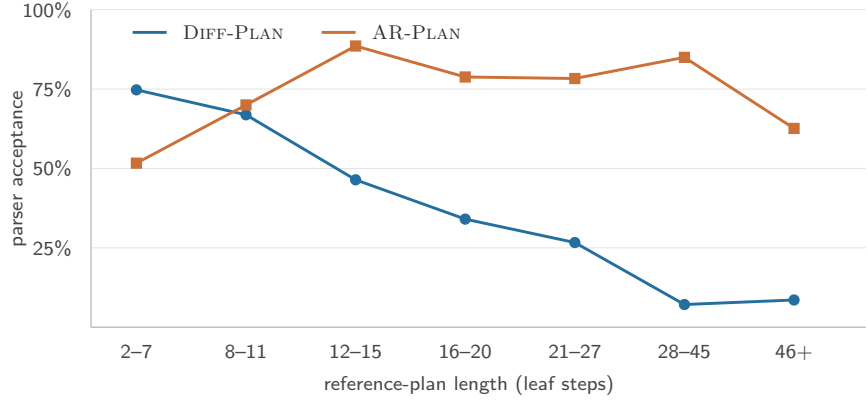
\begin{figure}[H]
\centering
\begin{tikzpicture}[
  x=1cm,y=1cm,
  axis/.style={gray!60,line width=0.45pt},
  grid/.style={gray!22,line width=0.35pt},
  dline/.style={diffcol,line width=1.1pt},
  arline/.style={arcol,line width=1.1pt},
  tick/.style={font=\sffamily\fontsize{7.5}{8.5}\selectfont,gray!55!black},
  legend/.style={font=\sffamily\fontsize{7.5}{8.5}\selectfont,gray!35!black},
]
\draw[axis] (0,0) -- (0,4.2);
\draw[axis] (0,0) -- (10.25,0);
\foreach \y/\lab in {1.05/25,2.10/50,3.15/75,4.20/100}{
  \draw[grid] (0,\y) -- (10.25,\y);
  \node[tick,anchor=east] at (-0.12,\y) {\lab\%};
}
\node[tick,rotate=90,anchor=south] at (-0.72,2.10) {parser acceptance};

\draw[dline] (0.60,3.14) -- (2.05,2.81) -- (3.50,1.95) -- (4.95,1.43) -- (6.40,1.12) -- (7.85,0.30) -- (9.30,0.36);
\foreach \x/\y in {0.60/3.14,2.05/2.81,3.50/1.95,4.95/1.43,6.40/1.12,7.85/0.30,9.30/0.36}{\filldraw[dline] (\x,\y) circle (0.055cm);}

\draw[arline] (0.60,2.17) -- (2.05,2.94) -- (3.50,3.72) -- (4.95,3.31) -- (6.40,3.29) -- (7.85,3.57) -- (9.30,2.63);
\foreach \x/\y in {0.60/2.17,2.05/2.94,3.50/3.72,4.95/3.31,6.40/3.29,7.85/3.57,9.30/2.63}{\filldraw[arline] (\x-0.055,\y-0.055) rectangle (\x+0.055,\y+0.055);}

\foreach \x/\lab in {0.60/{2--7},2.05/{8--11},3.50/{12--15},4.95/{16--20},6.40/{21--27},7.85/{28--45},9.30/{46+}}{
  \node[tick,anchor=north] at (\x,-0.10) {\lab};
}
\node[tick,anchor=north] at (5.0,-0.50) {reference-plan length (leaf steps)};

\draw[dline] (0.55,3.90) -- (1.00,3.90);
\node[legend,anchor=west] at (1.08,3.90) {\diffp};
\draw[arline] (3.05,3.90) -- (3.50,3.90);
\node[legend,anchor=west] at (3.58,3.90) {\arp};
\end{tikzpicture}
\caption{\textbf{ScienceWorld parser acceptance by reference-plan length.}
Rates are computed over $825$ held-out generations per planner. Acceptance for \diffp falls
from $74.7\%$ in the shortest bin to $7$--$9\%$ at $28$ or more reference steps,
with pooled parser acceptance of
$34.3\%$ for \diffp and $73.8\%$ for \arp.}
\label{fig:sw-length}
\end{figure}